\documentclass{article} % For LaTeX2e

\usepackage{iclr2026_conference,times}

\usepackage{amsmath,amsfonts,bm}

\def\eqref#1{equation~\ref{#1}}
\def\1{\bm{1}}

\DeclareMathAlphabet{\mathsfit}{\encodingdefault}{\sfdefault}{m}{sl}
\SetMathAlphabet{\mathsfit}{bold}{\encodingdefault}{\sfdefault}{bx}{n}

\usepackage{hyperref}
\usepackage{url}

\usepackage{booktabs}
\usepackage{makecell}
\usepackage{graphicx}
\usepackage{multirow}
\usepackage{xspace}
\usepackage{enumitem}
\usepackage{wrapfig}
\usepackage{adjustbox}
\usepackage{tabularx}
\usepackage{caption}
\usepackage{titletoc}
\usepackage{tikz}
\usepackage{bbding}
\usepackage{pifont}
\usepackage{setspace}
\usepackage[table]{xcolor}
\definecolor{LightBlue}{rgb}{0.88,1,1}
\iclrfinalcopy

\newcommand{\score}{\textsc{VideoScore2}\xspace}
\newcommand{\data}{\textsc{VideoFeedback2}\xspace}
\newcommand{\bench}{\textsc{VideoScore-Bench-v2}\xspace}

\title{\score: Think before You Score in Generative Video Evaluation}

\author{
$^{1}$Xuan He\thanks{\ \  Equal contribution. Xuan He is the project leader.}$^*$$^\dagger$ \ \ 
$^{2}$Dongfu Jiang$^*$$^\dagger$\ \  
$^{3}$Ping Nie \ \ 
$^{4}$Minghao Liu \ \
$^{7}$\textbf{Zhengxuan Jiang} \vspace{0.15em} \\ 
$^{2}$\textbf{Mingyi Su} \
$^{6}$\textbf{Wentao Ma} \ 
$^{6}$\textbf{Junru Lin} \ 
$^{2}$\textbf{Chun Ye} \ 
$^{6}$\textbf{Yi Lu} \
$^{2}$\textbf{Keming Wu} \vspace{0.15em} \\ 
$^{2}$\textbf{Benjamin Schneider} \
$^{2}$\textbf{Quy Duc Do} \
$^{2}$\textbf{Zhuofeng Li} \ 
$^{6}$\textbf{Yiming Jia} \
$^{2}$\textbf{Yuxuan Zhang} \vspace{0.15em} \\ 
$^{9}$\textbf{Guo Cheng} \
$^{2}$\textbf{Haozhe Wang} \
$^{5}$\textbf{Wangchunshu Zhou} \ 
$^{8}$\textbf{Qunshu Lin} \
$^{5}$\textbf{Yuanxing Zhang} \vspace{0.15em} \\
$^{2,5}$\textbf{Ge Zhang} \
$^{5}$\textbf{Wenhao Huang} \
$^{2}$\textbf{Wenhu Chen}$^\dagger$ \vspace{0.15em} \\
$^{1}$University of Illinois Urbana-Champaign, 
$^{2}$University of Waterloo,  \ 
$^{3}$Independent, \ 
$^{4}$2077AI,  \vspace{0.15em} \\
$^{5}$M-A-P,  \ 
$^{6}$University of Toronto,  \
$^{7}$Zhejiang University, \ 
$^{8}$Abaka AI, \
$^{9}$Netmind.AI
\vspace{0.4em}\\
$^\dagger$\texttt{xuanhe4@illinois.edu  \{dongfu.jiang, wenhuchen\}@uwaterloo.ca} 
}

\begin{document}

\maketitle

% \begin{abstract}
% Recent advances in text-to-video generation have produced increasingly realistic and diverse content, yet evaluating such videos remains a fundamental challenge. For existing evaluators or reward models, most are limited to a single score, lack interpretability, or provide only coarse analysis, making them insufficient for capturing the multi-faceted nature of video quality. We present \score, a multi-dimensional, interpretable, and human-aligned framework for assessing AI-generated videos. \score explicitly evaluates visual quality, text-to-video alignment, and physical/common-sense consistency, while also producing long-form rationales. The model is trained on large-scale human-annotated data with both scores and reasoning traces, starting with supervised fine-tuning for structured outputs and followed by reinforcement learning to enhance analytical robustness. Extensive experiments demonstrate that \score consistently achieves superior performance across both in-domain and out-of-domain evaluations. Beyond delivering reliable assessments, it also shows strong potential as a reward model for diffusion-based video generation, bridging the gap between evaluation and controllable generation.
% \end{abstract}

\begin{abstract}
Recent advances in text-to-video generation have produced increasingly realistic and diverse content, yet evaluating such videos remains a fundamental challenge due to their multi-faceted nature encompassing visual quality, semantic alignment, and physical consistency. Existing evaluators and reward models are limited to single opaque scores, lack interpretability, or provide only coarse analysis, making them insufficient for capturing the comprehensive nature of video quality assessment. We present \textsc{VideoScore2}, a \emph{multi-dimensional}, \emph{interpretable}, and \emph{human-aligned} framework that explicitly evaluates visual quality, text-to-video alignment, and physical/common-sense consistency while producing detailed chain-of-thought rationales. Our model is trained on a large-scale dataset \textsc{VideoFeedback2} containing 27,168 human-annotated videos with both scores and reasoning traces across three dimensions, using a two-stage pipeline of supervised fine-tuning followed by reinforcement learning with Group Relative Policy Optimization (GRPO) to enhance analytical robustness. Extensive experiments demonstrate that \textsc{VideoScore2} achieves superior performance with 44.35 (+5.94) accuracy on our in-domain benchmark \bench and 50.37 (+4.32) average performance across four out-of-domain benchmarks (VideoGenReward-Bench, VideoPhy2, etc), while providing interpretable assessments that bridge the gap between evaluation and controllable generation through effective reward modeling for Best-of-N sampling. Project Page: \url{https://TIGER-AI-Lab.github.io/VideoScore2/}.
\end{abstract}

\begin{figure}[ht]
    \vspace{-8pt}
    \centering
    \includegraphics[width=0.98\linewidth]{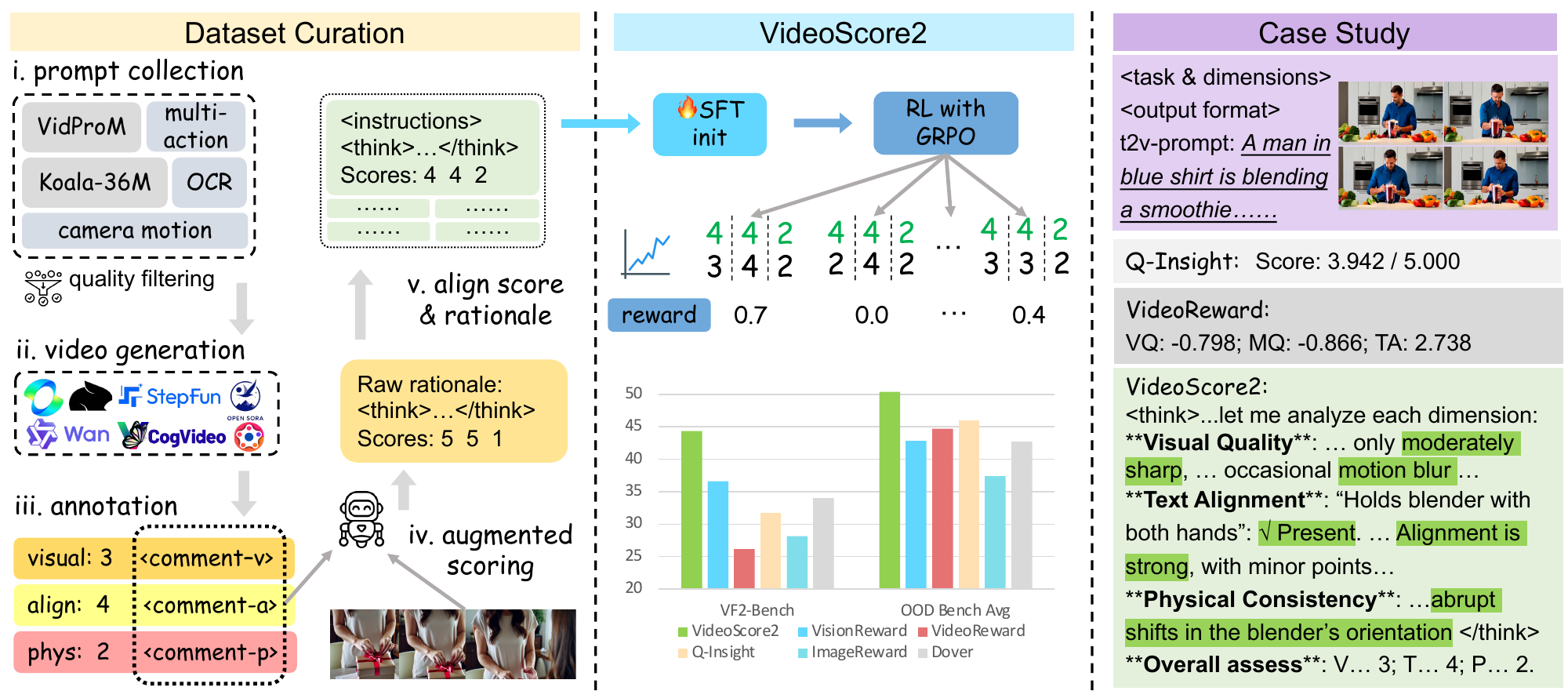}
    \caption{Overview of \score. We curate data from five different prompt sources and 22 T2V models with human annotated scores and rationale, which is further used for 2-stage training (cold-start SFT then RL) to elicit the model's thinking ability before scoring.}
    \label{fig:1_teaser}    
\end{figure}

\section{Introduction}
\label{sec:1}

Recent progress in text-to-video (T2V) generation~\citep{OpenAIsora,Kling-1.6,wan2025} has enabled models to produce increasingly realistic and coherent videos, expanding their potential across domains such as entertainment, education, and simulation. Yet, evaluating the quality of such videos remains a core bottleneck. Unlike text or image evaluation, video assessment must jointly consider visual fidelity, semantic adherence to prompts, and physical plausibility—dimensions. These abilities requires the backbone vision language model (VLM) to not only possess superior visual understanding ability but also comprehensive physical common senses.

Many powerful video evaluators/reward models have been developed these days and demonstrate strong performance in various video preference and point-score benchmarks. Examples works include VideoScore~\citep{he2024videoscorebuildingautomaticmetrics}, VideoPhy2~\citep{bansal2025videophy2challengingactioncentricphysical}, VideoReward~\citep{liu2025improving}, etc. However, while these works present superior ability in scoring accuracy, \textbf{they all collapse into a single opaque score, where no rationale is given for accountability}. What's more, these models are all trained via supervised fine-tuning on the collected dataset directly, limiting their generalization ability on the OOD data points~\citep{Chu2025SFTMR}.

To address this gap, we propose \score, a \emph{multi-dimensional}, \emph{interpretable}, and \emph{human-aligned} evaluator for AI-generated videos. \score not only outputs structured scores along three axes—\textbf{(1) visual quality}, \textbf{(2) text alignment}, and \textbf{(3) physical/common-sense consistency}—but also provides detailed chain-of-thought style analyses before giving its judgments. This “thinking-before-scoring” design makes \score unique among evaluators, enabling transparent and human-like reasoning. Furthermore, unlike some prior models whose SFT training restricts them to in-domain settings, \score demonstrates strong generalization across diverse out-of-domain benchmarks, confirming its robustness and reliability for video generation.

For training \score, we curated a large-scale dataset of multi-dimensional evaluations, \data, that combines quality scores with reasoning traces. 
The text-to-video prompts were sourced from both existing datasets and our manually designed cases in some special scenarios (e.g., multiple actions, OCR text, camera motion). The videos were generated by over twenty T2V models spanning from early baselines to recent state-of-the-art modern generative systems, marking a more fine-grained video quality gradient. Our annotators were instructed to provide scores (1-5) as well as brief diagnostic comments across three dimension, which were later expanded into detailed rationales through an LLM semi-blind scoring and alignment pipeline. This design yields a diverse, reliable, and reasoning-augmented dataset that serves as the foundation for teaching \score both what to evaluate and how to reason. As a result, we derived \textbf{2933 unique prompts}, \textbf{27168 generated videos} and \textbf{81504 scores with rationales} in total. We also split \textbf{500} exmaples as a new video point-score benchmark: \bench.

During the experiments, we adopt a two-stage training pipeline. First, a cold-start supervised fine-tuning (SFT) is applied to instill structured output formatting and basic reasoning capabilities. Then, we employ Group Relative Policy Optimization (GRPO) with RL through to further strengthen analytical robustness, refine interpretability, and align evaluations with human preference distributions. 
% This combination allows \score to balance format adherence, reasoning quality, and reward alignment.

Extensive experiments demonstrate the effectiveness of \score. On the in-domain \bench, \score achieves $44.35$ ($+5.94$) in point-score accuracy, $90.78$ ($+4.01$) in relaxed accuracy and $60.37$ ($+8.32$) in PLCC with significant improvements compared to the previous SoTA. Our model consistently achieves superior performance in the out-of-domain (OOD) benchmarks, reaching $50.37$ ($+4.32$) average performance across 2 OOD preference and 2 OOD point score benchmarks. Furthermore, we show \score's potential to be applied as a reward model for T2V generation via Best-of-N (BoN) sampling. We also conducted detailed ablation study to understand importance of rationale for SFT, cold-start SFT for RL, and score output format, etc. Results demonstrate RL with cold-start SFT and rationale as the best parctice.

\vspace{-6pt}
\section{Related Works}
\label{sec:2}
\vspace{-4pt}

\subsection{Text-to-Video Generation}
\label{subsec:2.1}
\vspace{-4pt}
Research on text-to-video (T2V) generation has progressed rapidly with the introduction of large diffusion and Transformer-based architectures. Early milestones include ModelScope \citep{wang2023modelscopetexttovideotechnicalreport}, which provided one of the first open-source diffusion pipelines for T2V, making the task widely accessible. Subsequent VideoCrafter2 \citep{chen2024videocrafter2} improved temporal fidelity with enhanced motion realism under data constraints. More recently, CogVideoX \citep{yang2024cogvideox} employed a large DiT backbone to achieve high resolution and narrative coherence. At the industrial scale, OpenAI Sora \citep{OpenAIsora} positions itself as a “world simulator,” capable of generating long videos with rich physical plausibility. Similarly, an open-sourced work StepVideo-T2V \citep{ma2025stepvideot2vtechnicalreportpractice} emphasizes scalable training and efficient architecture design to support long and coherent video synthesis. Other commercial systems such as Veo 3 \citep{Google-Veo3}, Kling-1.6 \citep{Kling-1.6}, and Pika-2.2 \citep{Pika2.2} further highlight advances in controllability, and human-centric generation. Despite these achievements, systematic and human-aligned evaluation of video qualities from visual perception to semantic reasoning remains limited, underscoring the need for multi-dimensional and interpretable evaluation frameworks.

\vspace{-6pt}
\subsection{Reward Modeling for Vision}
\label{subsec:2.2}
\vspace{-4pt}
Reward modeling has become a central paradigm for aligning generative models with human preferences in both image and video domains. Early methods such as Dover \citep{wu2023dover} and ImageReward \citep{xu2023imagereward} provide single scalar scores, which are effective but insufficient for capturing the multi-faceted nature of visual quality. More recent approaches—VideoReward \citep{liu2025improving}, UnifiedReward \citep{unifiedreward}, and Q-Insight \citep{li2025qinsight}—introduce multi-dimensional scoring, yet are limited to numeric ratings without explanatory reasoning. Other efforts like LiFT \citep{wang2025liftleveraginghumanfeedback} provide short analytical comments, but remain broad and lack the depth necessary for systematic evaluation. Addressing these limitations, \score delivers multi-dimensional assessments together with long-form analytical reasoning, making its evaluations both human-aligned and interpretable. Moreover, unlike many existing reward models whose reliance on SFT training often leads to poor generalization, \score demonstrates robust performance across OOD benchmarks, underscoring its potential as a more reliable evaluator.

\vspace{-6pt}
\subsection{Video Understanding and Reasoning}
\label{subsec:2.3}
\vspace{-4pt}

Video understanding and reasoning has been a long-standing problem in multimodal learning. Since 2022, transformer-based models have become the backbone of video understanding. Works like Video Swin Transformer \citep{liu2021videoswintransformer} and InternVideo \citep{wang2022internvideogeneralvideofoundation} show the benefit of large-scale pretraining and hierarchical temporal modeling. Extending to video–language reasoning, models such as Video-LLaMA \citep{zhang2023videollamainstructiontunedaudiovisuallanguage}, Video-LLaVA \citep{lin2024videollavalearningunitedvisual}, and mPLUG-Owl-V \citep{ye2024mplugowlmodularizationempowerslarge} align LMMs with video for open-ended QA and grounding. Recent efforts emphasize long video understanding, including LongVLM \citep{weng2024longvlmefficientlongvideo}, Video-ChatGPT \citep{Maaz2023VideoChatGPT}, and benchmarks like Video-MME \citep{fu2025videommefirstevercomprehensiveevaluation} and VideoEval-Pro \citep{ma2025videoevalprorobustrealisticlong}, which stress more realistic, open-ended evaluation of extended video reasoning.

\begin{table}[ht]
    \centering
    \begin{minipage}[ht]{0.54\textwidth}
        \small
        \centering
        \caption{Comparison of \score and existing reward models for multi-dimensions, rationale support, and dataset recency.}
        \label{tab:3_comparison}
        \renewcommand{\arraystretch}{1.2}
        \scalebox{0.8}{
        \begin{tabular}{@{}lcccc@{}}
\toprule
\textbf{Method} & \textbf{Input} & \textbf{Multi-Dim} & \textbf{Rationale} & \textbf{data recency (yy.mm)}  \\
\midrule
DeQA-Score  & image  & \ding{55}         & \ding{55}          & 24.10        \\
Q-Insight   & image  & \checkmark         & \ding{55}          & 24.10    \\
\midrule
VisionReward & video & \ding{55} & \ding{55}  & 24.08     \\
VideoReward  & video & \checkmark  & \ding{55}          & 24.09         \\

UnifiedReward  & video & \checkmark         & \ding{55}         & 24.12     \\
\midrule
\score  & video        & \checkmark         & \checkmark    & 25.04   \\   
\bottomrule
        \end{tabular}
        }
        \vspace{0.3em}
    \end{minipage}%
    \hfill
    \begin{minipage}[ht]{0.40\textwidth}
        \small
        \centering
        
        \includegraphics[width=0.85\linewidth]
        {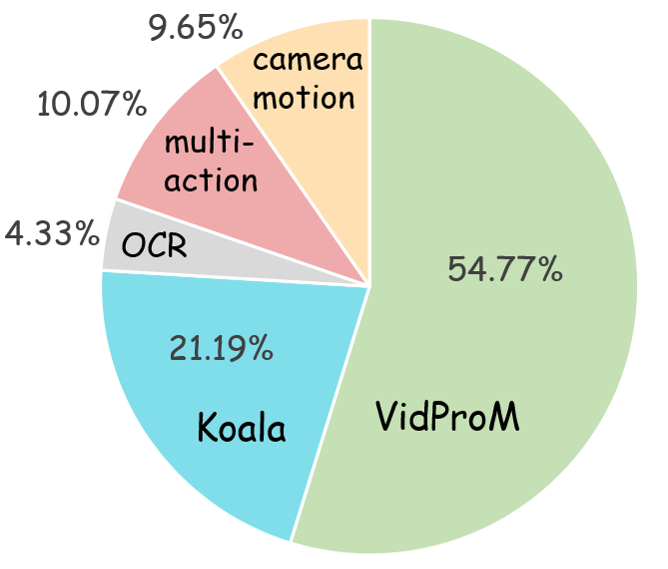}
        \captionof{figure}{Prompt source proportion.}
        \phantomcaption\label{fig:3_prompt_source}
        
        % \renewcommand{\arraystretch}{1.3}
        % \scalebox{0.88}{
        % \begin{tabularx}{0.98\linewidth}{l|ccc}
        % \toprule
        % Source & VidProM & \multicolumn{2}{c}{Koala-36M}  \\
        % \midrule
        % Prop. & 54.77\% & \multicolumn{2}{c}{21.19\%} \\
        % \midrule
        % Source & OCR-text & Multi-action &  Camera \\
        % \midrule
        % Prop. & 4.33\% & 10.07\% & 9.65\% \\
        % \bottomrule
        % \end{tabularx}
        % }
    \end{minipage}
\vspace{-8pt}
\end{table}

\vspace{-6pt}
\section{Dataset Curation}
\label{sec:3}
\vspace{-4pt}

\subsection{Data Preparation}
\label{subsec:3.1}
\paragraph{Prompt Collection.} 
\vspace{-4pt}

Our dataset prompts come from two sources: \textbf{existing datasets} VidProM \citep{wang2024vidprommillionscalerealpromptgallery} and Koala-36M \citep{wang2025koala36mlargescalevideodataset}, and \textbf{manually collected ones}. VidProM provides real user queries from generative model communities, while Koala-36M contains detailed and structured captions that can be adapted into text-to-video prompts. Since some raw prompts are abstract, incomplete or unsuitable, we adopt a two-stage filtering pipeline: (1) \textbf{rule-based filtering} is applied to remove prompts that are unsuitable due to length, format, or other constraints; (2) \textbf{LLM semantic filtering or revising} is used to discard or revise prompts that are abstract, incoherent, or bad for short video generation. See details
in Appendix~\ref{subsec:apdx_collecting_t2v_prompt}.

For the manually collected part, we focus on three categories: \emph{multi-action}, \emph{OCR text}, and \emph{camera motion}. This design is motivated by known limitations of current text-to-video (T2V) models in failing to express multiple actions~\citep{wang2024worldsimulatorgoodstory}, render readable text~\citep{}, and reproduce camera motion~\citep{}. To construct the multi-action and OCR text prompts, we first design about 100 seed examples and then ask LLMs to expand them creatively, while camera motion prompts are built by appending motion instructions (e.g., “pan left”, “tilt up”) to some sampled prompts directly.

\paragraph{Video Collection}
We collected videos from over 20 T2V models, ranging from early diffusion systems such as ModelScope \citep{wang2023modelscopetexttovideotechnicalreport} to advanced generators such as StepVideo-T2V \citep{ma2025stepvideot2vtechnicalreportpractice} and Kling-1.6 \citep{Kling-1.6}. For annotation, models were grouped into four coarse tiers (Poor/Early, Medium, Good, Perfect/Modern). For each prompt, we randomly sampled 10 models to generate videos, ensuring a balanced distribution across tiers. This design enabled direct comparisons among videos with the same semantic content but different quality levels, improving scoring consistency and reliability. By covering a wide range of resolutions (256$\times$256 to 1980$\times$982), frame rates (8–30 fps), and durations (1–6s), our dataset offers diverse variability, helping \score learn quality from poor output to near-photorealistic generations (see Appendix~\ref{subsec:apdx_video_stat} for details and ~\ref{subsec:apdx_video_examples} for video examples).

\paragraph{Evaluation Dimensions}
 We evaluate videos along three dimensions: \textbf{visual quality}, \textbf{text alignment}, and \textbf{physical / common-sense consistency}, to capture fidelity, semantic accuracy, and content-level reasoning. Unlike VideoScore with five dimensions\citep{he2024videoscorebuildingautomaticmetrics}, we remove dynamic degree (mostly prompt-dependent) and subsume temporal consistency under visual quality.

\begin{table}[!t]

\small
\centering
% \vspace{-4pt}
\vspace{-6pt}
\caption{Definition and Checklist of Evaluation Dimensions.}
\label{tab:3_eval_dims}
\renewcommand{\arraystretch}{1.2}
\scalebox{0.8}{
\begin{tabular}{ll}
\toprule
\textbf{Dimension} & \textbf{Definition and Checklist} \\ \midrule
Visual Quality  &  \makecell[l]{Quality of visual viewing experience, including resolution, overall and local \\ clarity, smoothness, brightness stability, distortions, etc.}\\
\midrule
Text Alignment  & \makecell[l]{The alignment between video content and text prompt, in terms of subjects, \\actions, details, styles, and sequential events, etc. } \\
\midrule
\makecell[l]{Physical / Common-sense \\ Consistency} &  \makecell[l]{Whether the video is normal and aligns with common sense, or physical laws, \\ based on everyday knowledge and intuition. Check for abrupt changes, \\ distortions, counterintuitive scenes, and anything weird and abnormal.} \\
\bottomrule 
\vspace{-8pt}
\end{tabular}
}
\end{table}

\vspace{-6pt}
\subsection{Annotation}
\label{subsec:3.2}
\vspace{-4pt}
We provide dimension-specific checklists in Table~\ref{tab:3_eval_dims} to help annotators understand the task. Annotators are required to assign \emph{integer scores} (1–5) and \emph{short comments} for each dimension, later expanded into full rationales by an LLM. For example, the comments can be: ``Low resolution, brightness is unstable'' or ``The second and third actions in prompt are missing''. Our team consists of 15 annotators who were trained with annotated examples and pilot rounds (30–50 videos each) with reviewer feedback to ensure consistency. See detailed guidelines in Appendx~\ref{subsec:apdx_anno_details}.

\begin{table}[!t]
    \centering
    \begin{minipage}[t]{0.47\textwidth}
        \small
        \centering
        \caption{Inter-Annotator Agreement (IAA) results ($\mathcal{R}$ / $\alpha$). $\mathcal{R}$ = Relaxed Match (all annotator scores within a margin of 1), $\alpha$ = Krippendorff's Alpha. }
        \label{tab:3_IAA}
        \renewcommand{\arraystretch}{1.3}
        \scalebox{0.88}{
        \begin{tabular}{@{}lccc@{}}
        \toprule
        \textbf{Trial} & \textbf{VQ} & \textbf{TA} & \textbf{PC} \\
        \midrule
        1 ($n=30$) &  93.33 / 92.06  & 93.33 / 82.71  &  83.33 / 82.99  \\
        2 ($n=30$) &  96.67 / 90.61 & 80.00 / 77.62  &  80.00 / 80.95 \\
        \bottomrule
        \end{tabular}
        }
        \vspace{0.3em}
    \end{minipage}%
    \hfill
    \begin{minipage}[t]{0.47\textwidth}
        \small
        \centering
        \renewcommand{\arraystretch}{1.3}
        \caption{Human inspection on the difference between model score and human score in augmented-scoring.}
        \label{tab:3_score_difference}
        \scalebox{0.88}{
        \begin{tabularx}{0.98\linewidth}{l|cccc}
        \toprule
        \multicolumn{5}{c}{\textbf{4951} videos, \textbf{14853} scores in total} \\
        \midrule
        \textbf{Difference}  & 0 & 1 & 2 & $\geq$3 \\
        \midrule 
        \textbf{Counts} & 4710 & 7698 & 2062 & 383 \\
        \midrule
        \multicolumn{5}{c}{\textbf{Bad videos} (diff. $\geq$3 in any dim.) : \textbf{337 / 4951}}   \\
        \bottomrule
        \end{tabularx}
        }
    \end{minipage}
\end{table}

Quality control was ensured through periodic audits, where 10–20\% of data was spot-checked for scoring accuracy and comment quality. Annotators with inconsistent work received feedback and were required to revise their annotations. As shown in Table~\ref{tab:3_IAA}, the inter-annotator agreement (IAA) indicates good labeling reliability.

\vspace{-6pt}
\subsection{SFT Data Processing }
\label{subsec:3.3}
\vspace{-4pt}

\paragraph{Rationale Elicitation}  
We use Claude-4-Sonnet \citep{claude-sonnet-4} (with thinking enabled) to elicit CoT-like rationales \citep{wei2023chainofthoughtpromptingelicitsreasoning}. The LLM receives evaluation instructions, sampled frames, annotator comments (without scores), and 2–3 few-shot examples (see Appendix~\ref{subsec:apdx_prompt_template}). Its outputs are compared with human scores, and reconciliation follows: (i) if the difference $\leq 1$, keep the human score; (ii) if the difference $=2$, average the two; (iii) if any dimension differs $\geq 3$, the whole entry is re-scored, up to three times. After resampling, fewer than 10\% of entries were discarded. In 4,951 videos ($14,853$ scores), we report the distribution of human–model differences in Table~\ref{tab:3_score_difference}.

% \begin{wrapfigure}{r}{0.45\textwidth}
%     \centering
%     \includegraphics[width=0.95\linewidth]{figures/3_thinking_len.png}
%     \caption{Rationale length (num of words). Most are in 200-600 words.}
%     \label{fig:3_thinking_len}
%     \vspace{-8pt}
% \end{wrapfigure}

\begin{figure}[!t]
    \vspace{-6pt}
    \centering
    \begin{minipage}{0.48\textwidth}
       \centering
        \includegraphics[width=0.95\linewidth]{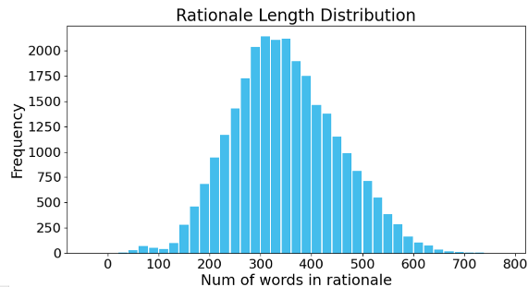}
        \caption{Rationale length (num of words). Most are in 200-600 words.}
        \label{fig:3_thinking_len}
    \end{minipage}
    \hfill
    \begin{minipage}{0.48\textwidth}
        \centering
        \includegraphics[width=0.95\linewidth]{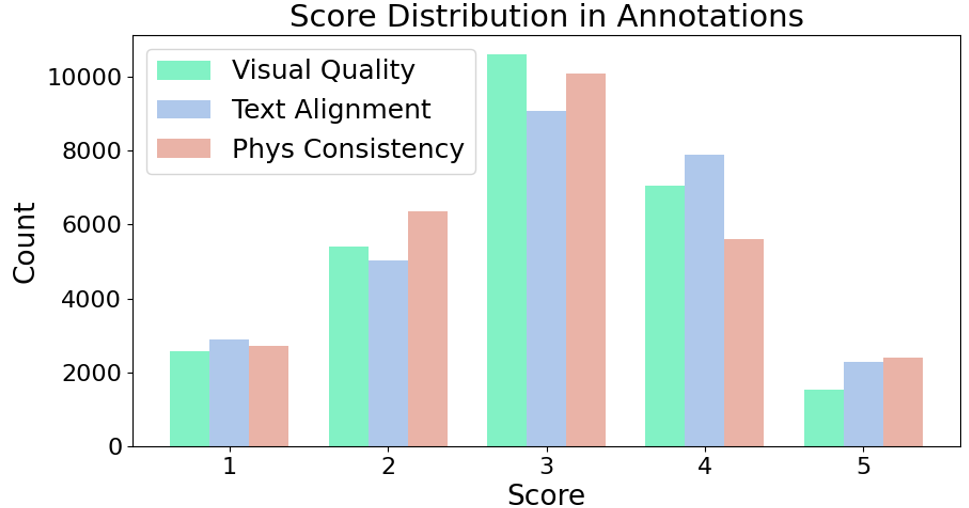}
        \caption{Human annotated score distribution in SFT data.}
        \label{fig:3_score_dist}
    \end{minipage}
    \vspace{-6pt}
\end{figure}

\paragraph{Align Rationales with Scores}  
Since the final score may occasionally differ from that mentioned in the rationale by one point, we use GPT-5-mini \citep{gpt-5} to align rationales with scores (prompt template shown in Appendix~\ref{subsec:apdx_prompt_template}). This lightweight adjustment preserved the rationale’s meaning while ensuring scoring consistency: typically, it involved only minor edits, such as softening or intensifying descriptions of quality issues (e.g., “slight blur’’ $\rightarrow$ “noticeable blur’’). The rationale length distribution is shown in Figure~\ref{fig:3_thinking_len}, and the score distribution in Figure~\ref{fig:3_score_dist}.

\paragraph{Building data for SFT and RL}
After processing, we obtain 27,168 samples (denoted as \data), and the proportions of videos across the four quality tiers (from best to worst) are 10.36\%, 33.53\%, 41.77\%, and 12.54\%, respectively (Appendix~\ref{subsec:apdx_video_stat}). 500 videos are held out as the test set (\bench) and the rest used for training. The SFT data follow a QA format, where the \emph{query} specifies the task (Table~\ref{tab:apdx_prompt_query_sft}), and the \emph{answer} provides rationale and scores. For RL, we follow Video-R1 \citep{liu2025improving}, using the same structure with \emph{problem} and \emph{solution} to compute accuracy rewards.

% \subsection{Statistics}
% \label{subsec:3.4}

\vspace{-4pt}
\section{VideoScore2}
\label{sec:4}
\vspace{-4pt}

% ============================== 4.1
\subsection{Training and Inference Setup}
\label{subsec:4.1}
\textbf{SFT Cold-Start.}  
We adopt a two-stage training strategy for \score. To ensure basic format-following ability and task familiarity, we first perform supervised fine-tuning (SFT) as the cold-start. The training is implemented with the LLaMA-Factory \citep{zheng2024llamafactory} framework, and Qwen2.5-VL-7B-Instruct \citep{Qwen2.5-VL} as base model.

For preparing the SFT checkpoint to initialize RL, we consider both the performance on \bench and training loss stability: high benchmark scores may indicate overfitting on in-domain tests and weak generalization to others. Balancing these factors, we adopt the configuration described in Appendix~\ref{subsec:apdx_sft_setup} as our main SFT model. Additional ablations on sampling fps, learning rate, and training epochs are reported in Appendix~\ref{subsec:apdx_ablation_sample_fps} and Appendix ~\ref{subsec:apdx_ablation_lr_epoch}.

% \textbf{Preliminaries on GRPO.} 
% GRPO is a lightweight variant of Proximal Policy Optimization (PPO) tailored for reinforcement learning with LLMs and VLMs. 
% Instead of relying on a value function, GRPO computes \emph{group-relative advantages} within a set of $G$ rollouts sampled for the same input query. 
% Concretely, given rollouts $\{y_i\}_{i=1}^G$ with rewards $\{R_i\}_{i=1}^G$, the group mean $\mu_g$ and variance $\sigma_g$ are used to normalize each reward:
% \begin{equation}
%     A_i = \frac{R_i - \mu_g}{\sigma_g},
% \end{equation} 
% This relative comparison stabilizes updates under sparse or coarse reward signals, making GRPO particularly suitable for our setting where evaluation scores are discrete and low-dimensional.

\textbf{Reinforcement Learning} We further train SFT checkpoint with open-source video reinforcement learning  framework Video-R1 \citep{feng2025videor1reinforcingvideoreasoning} implementing GRPO~\citep{shao2024deepseekmathpushinglimitsmathematical} to enhance its analytical robustness and human alignment:

\begin{itemize}[itemsep=-0pt,left=10pt,topsep=0pt]
    \item \emph{Accuracy Reward.}  
    The reward is defined by the degree of match. The design of this reward signal follows the principle that only predictions within $\pm 1$ of the ground truth on all dimensions should receive non-zero reward. On a 5-point scale, a deviation of one point is marginally acceptable, whereas deviations of two or more points indicate serious misjudgments and often contradict the ground-truth evaluation.
    \begin{equation}
        R_{\text{acc}} =
        \begin{cases}
            1.0 & \text{if all three dimensions match exactly}, \\
            0.7 & \text{if two match and one differs by 1}, \\
            0.4 & \text{if one matches and two differ by 1}, \\
            0.1 & \text{if all three differ by 1}, \\
            0   & \text{otherwise}.
        \end{cases}
    \end{equation}
    
    \item \emph{Format Reward.} 
    To ensure the output includes both a rationale and final scores, we assign $R_{\text{fmt}}=1$ if the response contains the \texttt{<think>} tag with rationale, and $R_{\text{fmt}}=0$ otherwise.
    
    \item \emph{Final Reward.} Following the setting for general video reasoning tasks in Video-R1, final reward $R = R_{\text{acc}} + \lambda R_{\text{fmt}}$. 
\end{itemize}

% \begin{wrapfigure}{r}{0.43\textwidth}
%     \centering
%     \includegraphics[width=0.96\linewidth]{figures/4_rl_acc_reward.png}
%     \caption{Accuracy rewards in RL training. }
%     \label{fig:3_thinking_len}
% \end{wrapfigure}

In practice, when starting from the SFT checkpoint, outputs already follow the query template, making the format reward redundant; thus we set $\lambda=0$ to focus on accuracy. In contrast, RL from the base model without SFT often shows format deviations, so we set $\lambda=0.3$ to encourage valid rationales and scores.  

RL training uses a learning rate of 2e-6 with $G=8$ generations per rollout, on 4$\times$A100 GPUs (about 8 hours per 100 steps). Evaluations on \bench and OOD benchmarks show performance peaks at around 300 steps; beyond this, performance on \bench drops (Appendix~\ref{subsec:apdx_rl_steps}). We therefore adopt the 300-step checkpoint for all reported results.

\textbf{Inference.}
Both the SFT and RL models output free-form text with rationale and final scores, following the same query template (Table~\ref{tab:apdx_prompt_query_sft}). To generalize discrete predictions $\{1,2,3,4,5\}$, we set decoding temperature to 0.7 and convert them into soft float scores using token-level probabilities:  
\begin{equation}
    \tilde{y} = \arg\max_{s} p(s) \times \frac{p(s)}{\sum_{j=1}^5 p(j)}.
\end{equation}
This yields smoother scores in $[1,5]$ while preserving interpretability. We further ablate score format (int vs. float) in Table~\ref{tab:4_ablation} and inference fps (2, 4, 8) in Appendix~\ref{subsec:apdx_ablation_infer}; all reported results use 2 fps and normalized float scores.

% ============================== 4.2 
\vspace{-6pt}
\subsection{Benchmarks} 
\label{subsec:4.2}
\vspace{-4pt}
In addition to the in-domain test on the \bench, we do further assessment on four out-of-domain benchmarks, testing the generalization ability across a wide range of video understanding and quality evaluation scenarios. The out-of-domain benchmarks can be categorized into two types based on the evaluation task: \textbf{Pairwise Preference} and \textbf{Point Score}. 

\textbf{Pairwise preference }benchmarks require the evaluator model to compare a pair of videos and identify which one exhibits higher quality.

\vspace{-4pt}
\begin{itemize}[itemsep=-0pt,left=10pt,topsep=0pt]
 \item \emph{VideoGenReward-Bench} \citep{liu2025improving}, built on VideoGen-Eval \citep{yang2025videogen}, contains 4,691 videos and 25,234 pairs. Annotators provide pairwise preference labels on dimensions of Visual Quality, Motion Quality, Text Alignment, and Overall preference.  

 \item \emph{T2VQA-DB} \citep{kou2024subjectivealigneddatasetmetrictexttovideo} assigns each video a human quality score (0–100). We sample 2,000 videos and derive 1,822 preference pairs by comparing the scores of videos.  
\end{itemize}
\vspace{-4pt}

The preference benchmarks both have ties, For models that output float scores, we treat two videos as having equal preference if their score difference is within 5\% of the model’s score range. (e.g., in $[0.0,5.0]$, scores 3.28 vs. 3.26 $\Rightarrow$ tie).

\textbf{Point score }benchmarks focus on how well the evaluator’s numerical predictions (after appropriate rescaling) align with the ground-truth scores in overall quality or fine-grained dimensions.

\vspace{-4pt}
\begin{itemize}[itemsep=-0pt,left=10pt,topsep=0pt]
 \item \emph{MJ-Bench-Video} \citep{tong2025mjvideofinegrainedbenchmarkingrewarding} contains 2,170 human-labeled videos with aspect-level scores in \{0,1,2\}. We use \{Fineness, Alignment, Coherence \& Consistency\} and average \score’s three dimensions to compare with their overall score.  

 \item \emph{VideoPhy2-test} \citep{bansal2025videophy2challengingactioncentricphysical} includes 3,396 videos annotated on a 1–5 scale for Semantic Adherence and Physical Consistency, which directly match \score’s second and third dimensions.  

 % \item \emph{AIGVE-Bench} \citep{xiang2025aigve} has 2,429 videos scored on a 1–5 scale across multiple dimensions. We align it with \score by either direct mapping or averaging related dimensions.  
\end{itemize}
\vspace{-4pt}

For consistency across benchmarks with different focuses and dimensions, we perform dimension mapping and ground-truth score rescaling, as detailed in Appendix \ref{subsec:apdx_modify_ood_bench}.

% ============================== 4.3 
\begin{figure}[!t]
    \centering
    \vspace{-6pt}
    \begin{minipage}[t]{0.48\textwidth}
        \centering
        \includegraphics[width=0.96\linewidth]{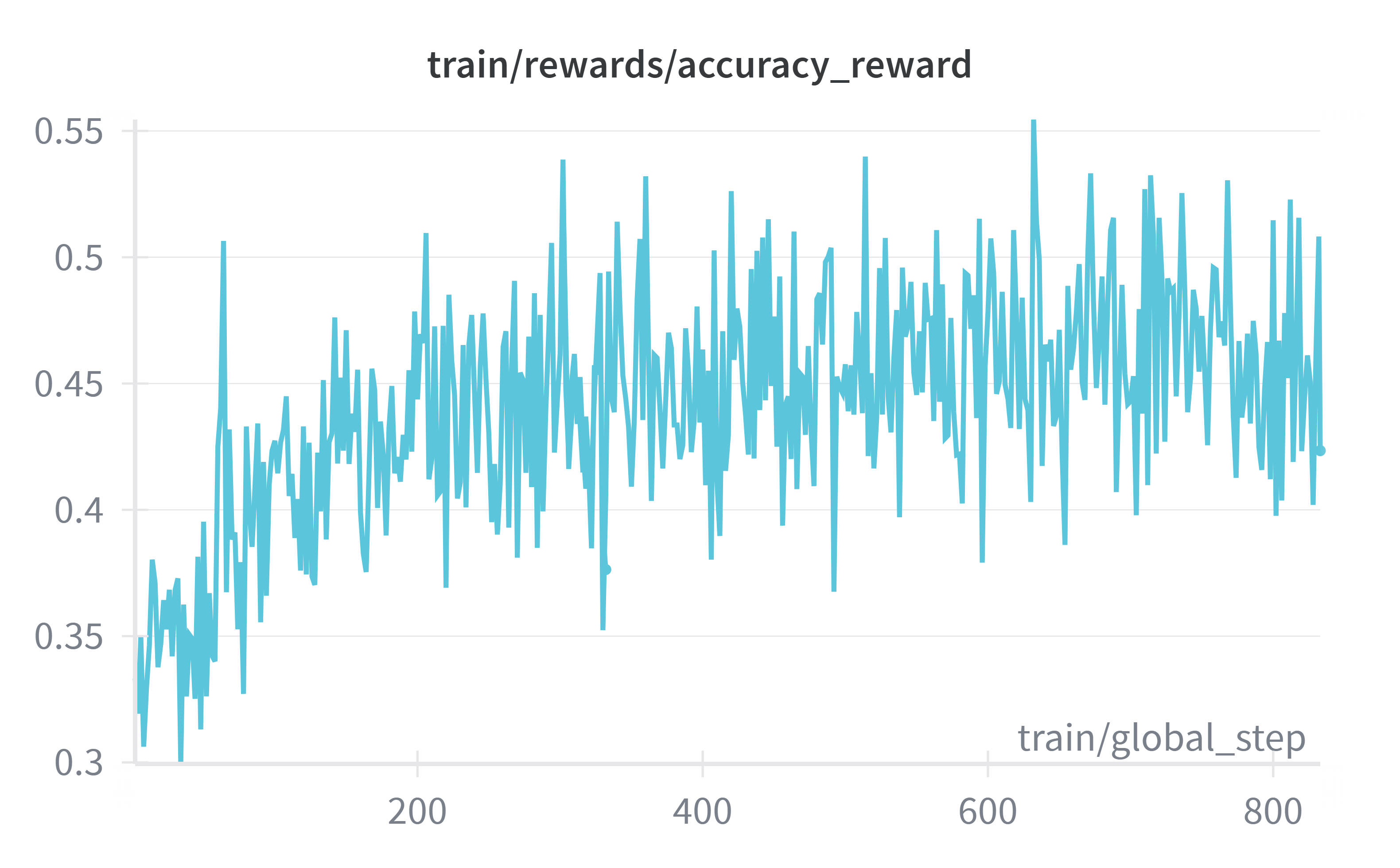}
        \caption{Accuracy rewards in RL training.}
        \label{fig:train_acc}
    \end{minipage}%
    \hfill
    \begin{minipage}[t]{0.48\textwidth}
        \centering
        \includegraphics[width=0.96\linewidth]{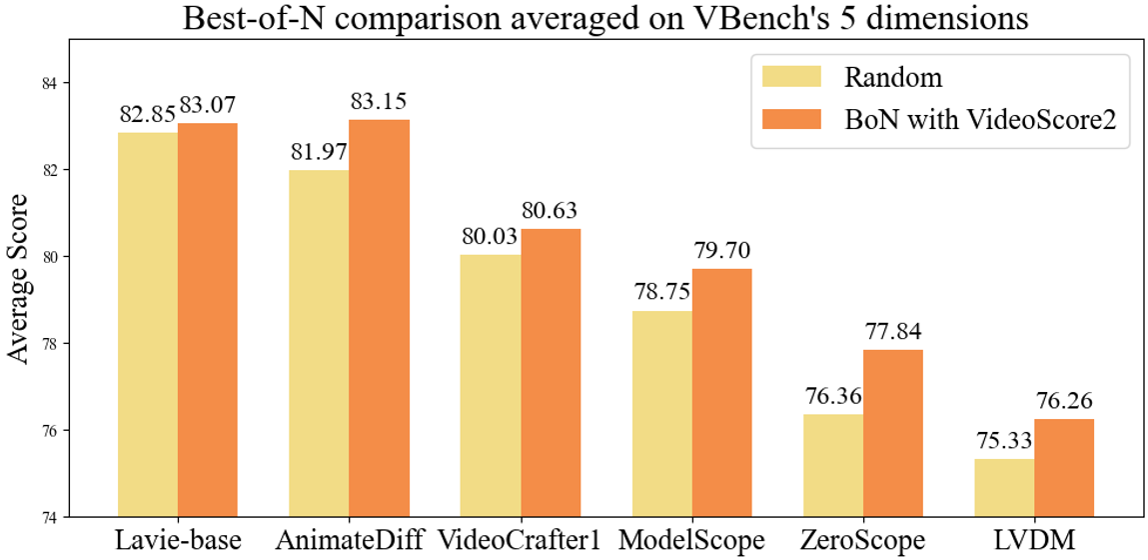}
        \caption{Comparison of Best-of-N sampling with \score and random ones on averaged 5 VBench dimensions.}
        \label{fig:4_BoN_hist}
    \end{minipage}
    \vspace{-6pt}
\end{figure}

\vspace{-6pt}
\subsection{Baseline Models}
\label{subsec:4.3}
\vspace{-4pt}
To ensure a comprehensive and rigorous evaluation of \score, we evaluate it against more than 10 baseline methods spanning diverse methodological families, divided into two classes.

\textbf{Prompting MLLMs}, we employ the same query templates as in the training data and provide sampled video frames as input to current most advanced LLMs with vision support for direct scoring, including Gemini-2.5-Pro \citep{gemini-2.5-pro}, GPT-5 \citep{gpt-5}, Claude-4-Sonnet \citep{claude-sonnet-4}, Grok-4 \citep{grok-4}, GLM-4.1v-9b-Thinking \citep{vteam2025glm45vglm41vthinkingversatilemultimodal}, Llama-4-Maverick \citep{llama-4} and Qwen2.5-VL-72B-Instruct \citep{Qwen2.5-VL}.

\textbf{Vision Reward/Scoring Models}, can be categorized based on whether it supports video input. 
\begin{itemize}[itemsep=-0pt,left=10pt,topsep=0pt]
    \item \textit{Image-only Models}:  
We adopt ImageReward~\citep{xu2023imagereward}, DeQA-Score~\citep{deqa_score}, and Q-Insight~\citep{li2025qinsight}. For video evaluation, frames are sampled at the same fps as the video models. ImageReward and DeQA-Score yield a single overall score, while Q-Insight supports aspect-specific queries, which we align with the three dimensions of \score.  
    \item \textit{Video-Capable Models}:  
We include VideoReward~\citep{liu2025improving}, UnifiedReward~\citep{unifiedreward}, VideoScore~\citep{he2024videoscorebuildingautomaticmetrics}, and VideoPhy2~\citep{bansal2025videophy2challengingactioncentricphysical}, which support multi-dimensional scoring. Others such as VisionReward~\citep{xu2024visionrewardfinegrainedmultidimensionalhuman}, Q-Align~\citep{wu2023qalign}, and DOVER~\citep{wu2023dover} provide only a single overall score.
\end{itemize}
Most models output scores without detailed reasoning. VisionReward uses fine-grained binary questions aggregated into a score, but lacks explicit explanations. LiFT adds short comments, yet these remain high-level and superficial. In contrast, \score produces both dimension-level scores and comprehensive analyses, making its evaluation more interpretable. Since different models use varying dimensions and scales, we rescale and adjust all outputs to match \score’s setting (detailed in  Appendix~\ref{subsec:apdx_baseline_rescaling}).

\begin{table}[!t]

\small
\centering
\caption{Accuracy and correlation between model answer and human score on \bench. \textit{Relaxed Accuracy} counts cases where the prediction differs from the ground truth by at most one point. \textbf{Bold} denotes the best model and the \underline{underlined} denotes the second best.}

% \scalebox{0.84}{
\resizebox{\textwidth}{!}{
\renewcommand{\arraystretch}{1.35}
\setlength{\tabcolsep}{5pt}
\begin{tabular}{l|ccc|c|ccc|c|ccc|c}
\toprule

\multirow{2}{*}{\textbf{\bench}} & \multicolumn{4}{c|}{Accuracy} & \multicolumn{4}{c|}{Relaxed Accuracy} & \multicolumn{4}{c}{PLCC}  \\  \cmidrule{2-13}
 & Visual  & Align  & Phy  & Avg  
 & Visual  & Align  & Phy  & Avg    
 & Visual  & Align  & Phy  & Avg  \\ 
 \midrule
% Random & 19.57                & 19.57                & 18.44                & 19.19                & 56.51                & 56.65                & 57.65                & 56.94                & -0.04                & 0.11                 & -1.28                &  -0.41  \\ 
% \midrule

\multicolumn{13}{c}{\textbf{Prompting MLLM}} \\ 
\midrule
Claude-Sonnet-4         & 33.07 & 29.86 & 23.85 & 28.93 & 76.35 & 76.95 & 61.92 & 71.74 & 20.17 & 30.64 & 18.01 & 22.94 \\
Gemini-2.5-Pro          & 29.92 & 29.72 & 24.07 & 27.90 & 71.49 & 70.88 & 61.45 & 67.94 & 26.71 & 32.96 & 19.75 & 26.47 \\
GPT-5                   & 30.72 & 27.91 & 20.08 & 26.24 & 73.90 & 72.29 & 60.84 & 69.01 & 13.38 & 23.34 & 17.24 & 17.99 \\
% Grok-4                  & 21.01 & 23.53 & 18.07 & 20.87 & 64.29 & 62.82 & 55.25 & 60.79 & 8.11  & 12.66 & 6.51  & 9.09  \\
% Qwen2.5-VL-72B & 28.84 & 27.91 & 19.07 & 25.27 & 72.33 & 77.21 & 55.58 & 68.37 & 14.51 & 6.63  & 0.85  & 7.33  \\
% Llama-4-Maverick        & 26.25 & 24.25 & 18.04 & 22.85 & 62.73 & 68.54 & 48.50 & 59.92 & 4.97  & 10.55 & 6.74  & 7.42  \\
GLM-4.1v-9B    & 33.27 & 31.46 & 21.42 & 28.72 & 80.76 & 77.15 & 61.22 & 73.04 & 28.03 & 16.80 & 10.18 & 18.34 \\
\midrule
\multicolumn{13}{c}{\textbf{Reward/Scoring Models for Image}} \\ \midrule
ImageReward             & 29.06 & 28.06 & 27.26 & 28.13 & 65.13 & 68.94 & 61.72 & 65.26 & 28.23 & 40.76 & 23.26 & 30.75 \\
DeQA-Score                    & 36.87 & 28.66 & 32.06 & 32.53 & 85.37 & 77.15 & 80.96 & 81.16 & 44.87 & 23.96 & 29.73 & 32.85 \\
Q-Insight               & 33.60 & 30.60 & 31.00 & 31.73 & 81.40 & 77.40 & 75.60 & 78.13 & 41.05 & 25.44 & 27.54 & 31.34 \\
\midrule
\multicolumn{13}{c}{\textbf{Reward/Scoring Models for Video}} \\ 
\midrule
VideoScore1.1           & \underline{41.48} & \underline{34.87} & \underline{38.88} & \underline{38.41} & \underline{90.98} & 82.37 & \underline{86.97} & \underline{86.77} & 49.00 & 30.90 & 47.00 & 42.30 \\
VideoReward             & 23.45 & 28.86 & -     & 26.16 & 60.32 & 67.74 & -     & 64.03 & 46.36 & \underline{48.31} & -     & 47.34 \\
UnifiedReward           & 25.20 & 27.20 & 22.80 & 25.07 & 71.00 & 64.80 & 68.00 & 67.93 & \underline{58.61} & 43.91 & \textbf{53.64} & \underline{52.05} \\
VisionReward            & 41.28 & 33.47 & 35.07 & 36.61 & 87.17 & \underline{84.37} & 82.16 & 84.57 & 46.85 & 45.32 & 38.25 & 43.47 \\
Q-Align                 & 28.66 & 28.06 & 27.86 & 28.19 & 75.55 & 69.74 & 68.94 & 71.41 & 54.71 & 34.01 & 37.78 & 42.17 \\
AIGVE-MACS              & 20.12 & 12.48 & 14.09 & 15.56 & 62.37 & 46.48 & 45.27 & 51.37 & 27.30 & 6.90  & 13.03 & 15.74 \\
VideoPhy2-AutoEval      & -     & 28.46 & 16.23 & 22.35 & -     & 73.75 & 52.31 & 63.03 & -     & 35.42 & 25.41 & 30.42 \\
Dover                   & 39.08 & 31.06 & 31.86 & 34.00 & 84.77 & 74.75 & 75.92 & 78.48 & 50.24 & 32.83 & 33.00 & 38.69 \\
\midrule
\multicolumn{13}{c}{\textbf{\score}} \\ 
\midrule
% Ours (SFT only) & 43.69 & 40.88 & 34.87 & 39.81 & 90.38 & 86.97 & 83.77 & 87.04 & 56.74 & 58.24 & 44.72 & 53.23 \\

% Ours (RL w/o SFT) & 40.12 & 35.09 & 34.89 & 36.70 & 84.18 & 81.54 & 83.37 & 83.03 & 53.43 & 59.92 & 46.93 & 53.43   \\

% Ours (RL + SFT) & \textbf{50.10} & \textbf{43.88} & \textbf{39.08} & \textbf{44.35} & \textbf{92.99} & \textbf{91.38} & \textbf{87.98} & \textbf{90.78} & \textbf{60.13} & \textbf{62.60} & \underline{52.73} & \textbf{60.37}   \\

Ours & \textbf{50.10} & \textbf{43.88} & \textbf{39.08} & \textbf{44.35} & \textbf{92.99} & \textbf{91.38} & \textbf{87.98} & \textbf{90.78} & \textbf{60.13} & \textbf{62.60} & \underline{52.73} & \textbf{60.37}   \\

\rowcolor{LightBlue} $\Delta$ over Best Baseline & +8.62 & +9.01 & +0.20 & +5.94 & +2.80 & +7.01 & +1.01 & +4.01 & +1.52 & +14.29 & -0.91 & +8.32  \\
\bottomrule
\end{tabular}
}
\vspace{-4pt}

\label{tab:4_vs2_bench}
\end{table}
\renewcommand{\arraystretch}{1.25}
\begin{table}[!ht]
\small
\centering
% \vspace{-0.5em}
\caption{Performance comparison on out-of-domain benchmarks, with 2 pairwise preference benchmarks and 3 point-score benchmarks. \textbf{Bold} denotes the best model and the \underline{underlined} denotes the second best. For “OOD Preference Benchmark,” performance is computed over all test samples.}
\scalebox{0.9}{
% \resizebox{\textwidth}{!}{
\setlength{\tabcolsep}{5pt}

\begin{tabular}{l|c|cc|cc}
\toprule
 \multirow{3}{*}{\textbf{OOD Bench}} & \multirow{3}{*}{\textbf{Average}}  & \multicolumn{2}{c|}{\textbf{OOD Preference Benchmark}}  & \multicolumn{2}{c}{\textbf{OOD Point Score Benchmark}}  
 \\ 
\cmidrule(lr){3-4} \cmidrule(lr){5-6} 
 
 &  & \textbf{VideoGen-}    & \textbf{T2VQA-DB}     & \textbf{MJ-Bench}    & \textbf{VideoPhy2}     \\

 & & \textbf{Reward Bench}  & \textbf{(Preference)} & \textbf{-Video}    & \textbf{-test}       \\ 
\midrule

\multicolumn{6}{c}{\textbf{Reward/Scoring Models for Image}} \\ 
\midrule

ImageReward & 37.40 & 47.14 & 43.46 & 37.51 & 21.48 \\
DeQA-Score  & 40.54 & 53.88 & 35.22 & 44.19 & 28.85  \\
Q-Insight   & \underline{46.05} & 54.05 & 46.65 & 52.58 & 30.90  \\
\midrule

\multicolumn{6}{c}{\textbf{Reward/Scoring Models for Video}} \\ 
\midrule

VideoScore-v1.1 & 38.87 & 16.79 & 39.18 & \textbf{71.57} & 27.95  \\
VideoReward     & 44.73 & \textbf{59.69} & 36.15 & 51.75 & 31.33  \\
UnifiedReward   & 37.22 & 53.31 & \underline{50.39} & 23.18 & 22.02  \\
VisionReward    & 42.86 & \underline{54.31} & 37.64 & 56.91 & 22.58  \\
Q-Align         & 32.62 & 42.05 & 43.24 & 21.97 & 23.22  \\
AIGVE-MACS      & 30.48 & 37.09 & 36.91 & 31.00 & 16.93  \\
VideoPhy2       & 29.13 & 30.75 & 24.12 & 24.00 & \textbf{37.64} \\
Dover           & 42.70 & 54.27 & 44.62 & 43.69 & 28.21  \\
\midrule

\multicolumn{6}{c}{\textbf{\score}} \\ 
\midrule

% Ours (SFT only)   & 46.98 & 50.79 & 52.36 & \underline{66.88} & 30.02 & 34.83 \\
% Ours (RL w/o SFT) & 43.14 & \underline{54.53} & \textbf{54.54} & 56.43 & 27.69 & 22.53 \\
% Ours (SFT + RL)   & \textbf{47.44} & 51.53 & 50.60 & 65.77 & \underline{33.58} & 35.74 \\

Ours   & \textbf{50.37} & 51.53 & \textbf{50.60} & \underline{65.77} & \underline{33.58}  \\

\bottomrule

\end{tabular}
}

\label{tab:4_ood_bench}
\end{table}

% ============================== 4.4
\vspace{-6pt}
\subsection{Evaluation Results}
\label{subsec:4.4}
\vspace{-6pt}
We report results on \bench in Table~\ref{tab:4_vs2_bench}, with Accuracy (w/o and w/ relaxation) and correlation metrics (PLCC). For float-output models, scores are rounded for accuracy and kept raw for correlations. \score surpasses the best baseline across all dimensions and metrics. We further test on four out-of-domain (OOD) benchmarks: two pairwise preference and two point-score (Section~\ref{subsec:4.2}). In the tables, \textit{Overall} denotes an explicit overall score, while \textit{Avg} is the mean across dimensions; preference results include ties. As shown in Table~\ref{tab:4_ood_bench}, while \score is not always the top model on each benchmark, it achieves the highest overall average.

To further validate the effectiveness of \score in video evaluation, we conduct human inspection to examine whether its predicted scores were reasonable and whether the analyses were accurate and appropriate. Qualitative examples are provided in Appendix~\ref{sec:apdx_case_study}.

% \begin{wrapfigure}{r}{0.5\textwidth}
%     \centering
%     \includegraphics[width=0.96\linewidth]{figures/4_BoN_hist.png}
%     \caption{Comparison of Best-of-N sampling with \score and random ones on averaged 5 VBench dimensions.}
%     \label{fig:4_BoN_hist}
% \end{wrapfigure}

% ============================== 4.5 
\vspace{-6pt}
\subsection{Best-of-N Sampling with \score}
\label{subsec:4.5}
\vspace{-6pt}
We evaluate \score with best-of-$n$ (BoN) sampling ($n=5$), where the model selects the best video among candidates. Six T2V models of moderate or poor quality are used, avoiding very strong ones to highlight the BoN effect. For 500 prompts, each model generates $500 \times 5$ videos.  
Comparison on VBench (Figure~\ref{fig:4_BoN_hist}) shows BoN consistently outperforms random sampling, confirming that \score effectively guides higher-quality selection. See full results in Appendix~\ref{subsec:apdx_full_res_BoN}.

% \input{tables/4_best_of_n}

% \begin{figure}[htbp]
%     \centering
%     \includegraphics[width=0.68\textwidth]{figures/4_BoN_hist.png}
%     \caption{Best-of-N results. Comparison of sampling with \score and random ones on 5 VBench dimensions, averaged score are shown. BoN with \score demonstrates consistent improvement.}
%     \label{fig:4_BoN_hist}
% \end{figure}

% ============================== 4.6
\vspace{-6pt}
\subsection{Ablation Study}
\label{subsec:4.6}
\vspace{-6pt}
\begin{table}[!t]

\small
\centering
% \vspace{-4pt}
\caption{Ablations on RL start point, rationale in SFT and score output format.}
% \scalebox{0.9}{
\resizebox{\textwidth}{!}{
\renewcommand{\arraystretch}{1.25}
\setlength{\tabcolsep}{5pt}
\begin{tabular}{l|c|cc|cc}
\toprule
 \multirow{4}{*}{\textbf{Ablations}} & \textbf{In-Domain} & \multicolumn{2}{c|}{\textbf{OOD Preference Benchmark}} & \multicolumn{2}{c}{\textbf{OOD Point Score Benchmark}} \\

\cmidrule(lr){2-6}

\multirow{3}{*}{} & \textbf{\bench} & \textbf{VideoGen-}      & \textbf{T2VQA-DB}      & \textbf{MJ-Bench} & \textbf{VideoPhy2}  \\
&  & \textbf{Reward-Bench} &  \textbf{(Preference)} & \textbf{-Video} & \textbf{-test}  \\

\midrule
% \multicolumn{6}{c}{\textbf{RL start w/ and w/o SFT}} \\ 
% \midrule

 RL w/o SFT  & 36.70 & \textbf{54.53}  & \textbf{54.54}  & 56.43 & 27.69   \\
 
 RL w/ SFT  & \textbf{44.53} & 51.53  & 50.60  & \textbf{65.77} & \textbf{33.58}   \\

\midrule
% \multicolumn{6}{c}{\textbf{SFT w/ and w/o CoT-like rationale}} \\ 
% \midrule

w/ CoT (default) & \textbf{39.81} & 50.79  & 52.36  & \textbf{66.88} & \textbf{30.02}  \\

w/o CoT & 32.17 & \textbf{54.74}   & \textbf{58.63}  & 59.06 & 21.83    \\

 \midrule
% \multicolumn{6}{c}{\textbf{Score output format}} \\ 
% \midrule

 Normalized (default) & 44.53 & \textbf{51.53}  & \textbf{50.60}  & 65.77 & 33.58   \\
 
 Raw Int Score  & \textbf{45.83} & 51.19 & 30.22 & \textbf{66.51} & \textbf{34.51}   \\
 
\bottomrule
\end{tabular}

}
\vspace{-4pt}

\label{tab:4_ablation}
\end{table}
Besides the ablations on SFT settings, RL training steps, as well as inference configurations (Appendix ~\ref{sec:apdx_exp_and_ablation}), we conduct the following studies, providing more insights of designing \score, summarized in Table~\ref{tab:4_ablation}.

\textbf{Cold Start.}  
We compare RL initialized from the base Qwen2.5-VL-7B-Instruct versus the SFT checkpoint. The SFT version achieves higher average scores across both \bench\ and OOD benchmarks, even if not superior on every benchmark. This indicates SFT provides a stronger starting point, enabling RL to focus on reward alignment rather than task formatting.  

\textbf{SFT w/ and w/o rationale} We further test SFT with and without CoT-like rationales. While the CoT-based version is slightly weaker on preference benchmarks, it performs significantly better on point-score benchmarks and thus improves generalization on average. This confirms that rationales are not only important for interpretability but also beneficial for overall robustness. 

\textbf{Score format.}  
We ablate the output format by comparing raw integer scores and normalized float scores. While integers show slight advantages on OOD point-score benchmarks, they perform notably worse on OOD preference tasks. Using normalized float scores strikes a better balance, preserving accuracy for point-score while capturing finer quality differences in preference settings.

\section{Conclusion}
\label{sec:5}
\vspace{-4pt}
In this work, we introduced \score for multi-dimensional, interpretable, and human-aligned evaluation of AI-generated videos. By building a comprehensive annotation pipeline that gathers diverse prompts, generative videos as well as reliable scores and rationales, we are able to train \score in the 2-stage paradigm. Comprehensive experiments demonstrate that our model outperforms existing evaluators across in-domain and out-of-domain benchmarks. We believe that \score open a path for trustworthy evaluation and human-aligned training of generative video models. Furthermore, our evaluation results also shows that model still struggle in evaluating physics and common senses in the generative models, highlighting the importance of a world model for video evaluator. We leave this as a future direction worth to explore.

\section*{Ethics Statement}
This work adheres to the ICLR Code of Ethics. In this study, no human subjects or animal experimentation was involved. All datasets used, including our curated \data, were sourced and processed in compliance with relevant usage guidelines, ensuring no violation of privacy or intellectual property.  

For prompt collection, we applied strict filtering to exclude NSFW, harmful, or otherwise inappropriate content, and ensured that prompts did not involve personally identifiable information (PII) or sensitive entities. Videos used for annotation were generated by publicly available text-to-video models, and only non-sensitive, safe prompts were retained. Our annotation guidelines emphasized fairness and consistency, and all annotators were trained to avoid introducing biased or discriminatory judgments.  

No personally identifiable information was collected or used, and no experiments were conducted that could raise privacy or security concerns. We are committed to maintaining transparency, fairness, and integrity throughout the research process.

\section*{Reproducibility statement}
All the code and datasets used in the paper will be open-sourced after the paper is accepted. We also have provided comprehensive details for both training (see in Appendix~\ref{sec:apdx_exp_and_ablation}) and evaluation (see in Appendix~\ref{sec:apdx_full_res}) to help the community for reproduction.

\bibliography{iclr2026_conference}
\bibliographystyle{plainnat}

\newpage
\appendix
\label{sec:appendix}
\section*{Appendix}
\startcontents[sections]
\printcontents[sections]{l}{1}{\setcounter{tocdepth}{3}}
\clearpage
\section{Data Collection and Processing}

% =======================
\subsection{Collecting Text-to-Video Prompts}
\label{subsec:apdx_collecting_t2v_prompt}
\subsubsection*{Source 1: VidProM} 

\paragraph{Rule-based filtering.} 

\begin{itemize}
    \item \emph{NSFW probability.} The original dataset provides a probability that a given prompt may lead to NSFW content. We exclude prompts with NSFW probability greater than 0.2, following the original dataset’s setting.  
    
    \item \emph{Trigger-word filtering.}  
    Exclude prompts intended for \emph{image-to-video} generation, which explicitly mention image attachments, and prompts specifying aspect ratios, or duration, which cannot be freely controlled in most T2V models. The trigger-word list includes: \texttt{["screen size", "16:9", "1:1", "3:4", "4k", "8k", "seconds", "message", "attach"]}.

    \item \emph{Length control.} Only prompts between 15 and 100 words are retained.  
    
\end{itemize}

\paragraph{LLM Semantic filtering.}
To filter out unsuitable prompts, we use GPT-4o-mini for semantic checks and exclude problematic ones. Specifically, we remove prompts that:

\begin{itemize}
\item vague or meaningless, lacking a concrete task,  
\item containing specific people or names,  
\item missing substantive verbs or motion, closer to images than videos,  
\item describing over three actions or events, too complex for short videos.  
\end{itemize}

\vspace{0.5\baselineskip}

\subsubsection*{Source 2: Koala-36M} 
\paragraph{Rule-based filtering.}
\begin{itemize}
    \item Since prompts come from real video captions, we only keep those associated with video segments shorter than 5 seconds; longer captions usually describe too many actions and are unsuitable for short video generation.  
    \item Each video–caption pair includes a \emph{clarity score} and an \emph{aesthetic quality score}. We exclude captions with clarity score below 0.95 or aesthetic score below 4.0. 
\end{itemize}

\paragraph{LLM Semantic filtering and revising} 
\begin{itemize}
    \item Same semantic checks as for VidProM, removing ambiguous or low-quality prompts.  
 
\end{itemize}

\vspace{0.5\baselineskip}

\subsubsection*{Source 3: OCR-Text (manually collected).}
For the OCR-text category, we first drafted seed prompts that \textbf{explicitly required text to appear} in the video, then expanded them using LLMs to create realistic yet creative scenarios where text naturally integrates into the scene. These prompts are diverse and challenging, often harder to generate than purely human-written ones. For example:
\begin{itemize}
    \item A painter adds brush strokes to a canvas, with a palette that says `Portrait of a Lady, Acrylic Paints, Warm Tones and Fine Detail'.  
    \item A photographer adjusts their lens with `Capture the Perfect Shot: Photography Tips and Tricks' displayed on a screen in front of them.  
\end{itemize}
In total, 200 prompts were collected.  

\vspace{0.5\baselineskip}

\subsubsection*{Source 4: Multi-Action (manually collected).}

For the multi-action category, we followed a similar approach as OCR-text. We first drafted seed prompts containing \textbf{two or three connected actions}, then expanded and rewritten them with LLMs to produce diverse, story-like scenarios. In total, 200 prompts were collected, each describing a short narrative with three consecutive actions. For example:
\begin{itemize}
    \item A woman adjusts her glasses, glances at the book with focus, and flips to the next page with a smile.  
    \item A fluffy orange cat swats a ball of yarn, sends it rolling, then dashes after it and pounces mid-roll.  
\end{itemize}

\vspace{0.5\baselineskip}

\subsubsection*{Source 5: Camera Motion (manually collected).}

For the camera-motion category, we did not generate entirely new prompts. Instead, we augmented existing prompts by appending explicit camera movement instructions at the end. Common motions include \textit{``Zoom in,'' ``Zoom out,'' ``Pan left,'' ``Pan right,'' ``Pan up,'' ``Pan down,'' ``Tilt up,'' ``Tilt down,''} and \textit{``Tracking shot.''} This simple yet effective strategy allows the dataset to capture scenarios where video realism depends on both content generation and dynamic camera behavior.

\begin{figure}[ht]
    \centering
    \begin{minipage}{0.48\textwidth}
       \centering
        \includegraphics[width=0.95\linewidth]{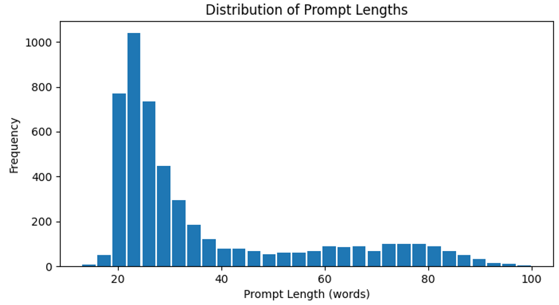}
        \caption{Distribution of Length (Num of Words) for the Prompt Set.}
        \label{fig:apdx_prompt_len_hist}
    \end{minipage}
    \hfill
    \begin{minipage}{0.48\textwidth}
        \centering
        \includegraphics[width=0.95\linewidth]{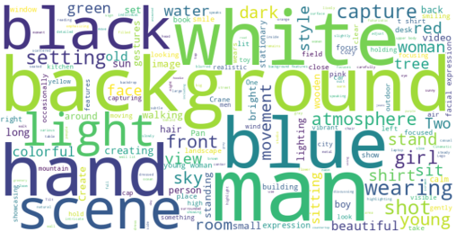}
        \caption{Word Cloud for the Prompt Set.}
        \label{fig:apdx_prompt_word_cloud}
    \end{minipage}
\end{figure}

% =======================
\clearpage
\subsection{Statistics of Generated Videos}
\label{subsec:apdx_video_stat}
We generate videos for annotation using more than twenty text-to-video (T2V) models, spanning from early diffusion-based systems such as ModelScope \citep{wang2023modelscopetexttovideotechnicalreport} to recent high quality generators like Kling-1.6 \citep{Kling-1.6}. This ensures a broad quality spectrum, covering both weak and strong generations. 

As discussed in Section~\ref{subsec:3.1}, to facilitate fairer comparisons and improve annotation reliability, we categorize these models into four coarse quality tiers. For each prompt, ten videos are sampled from ten different models while maintaining a balanced distribution across the four tiers. Typically 1-2 from "Poor / Early", 3-4 from Moderate, 3-4 from "Good", and 1 from "Perfect / Modern". The resulting videos vary widely in characteristics, with durations ranging from 1 to 6 seconds, resolutions from 256$\times$256 up to 1920$\times$982, and frame rates from 8 to 30 fps. A full summary of the models (and its variants) used is provided in Table~\ref{tab:apdx_video_stat}.

\begin{table}[!h]
\small
\centering
\caption{Detailed information of videos in our dataset, including t2v-model sources, video fps, resolution, duration, etc.}
\scalebox{0.92}{
% \resizebox{\textwidth}{!}{
\renewcommand{\arraystretch}{1.4}
\setlength{\tabcolsep}{4pt}
\begin{tabular}{l|c|ccc|cc}
\toprule
T2V Model (Suffix code in dataset) & Open Source & FPS & Resolution & Duration & Num & Proportion \\
\midrule
\multicolumn{7}{c}{\textbf{Tier1: Perfert / Modern.} 2814 videos, 10.36\%} \\ 
\midrule
Kling-1.6 \citep{Kling-1.6} (r) & N & 24.0 & 1280*720 & 5.0s & 611 & 2.25\% \\
Sora \citep{OpenAIsora} (s) & N & 30.0 & 1920*982 & 10.0s & 298 & 1.10\% \\
Pika-2.2 \citep{Pika2.2} (t) & N & 24.0 & 1280*720 & 5.0s & 321 & 1.18\% \\
StepVideo-T2V \citep{ma2025stepvideot2vtechnicalreportpractice} (y) & Y & 25.0 & 992*544 & 4.0s & 741 & 2.73\% \\
Wanx-2.1 (14B) \citep{wan2025} (w) & Y & 25.0 & 832*480 & 3.9s & 281 & 1.03\% \\
Ruyi \citep{createai2024ruyi} (A) & Y & 24.0 & 1008*576 & 5.0s & 184 & 0.68\% \\
CogVideoX-1.5 \citep{yang2024cogvideox} (g) & Y & 10.0 & 1360*768 & 4.0s & 378 & 1.39\% \\
\midrule
\multicolumn{7}{c}{\textbf{Tier2: Good.} 9598 videos, 33.53\%} \\ 
\midrule
Wanx-2.1 (1.3B) \citep{wan2025} (v) & Y & 24.0 & 832*480 & 3.9s & 1497 & 5.51\% \\
MagicTime \citep{yuan2025magictime} (q) & Y & 8.0 & 512*512 & 2.0s & 1741 & 6.41\% \\
Mochi1-Preview \citep{genmo2024mochi} (c) & Y & 10.0 & 848*480 & 1.9s & 1649 & 6.07\% \\
LaVie-base \citep{wang2023lavie} (h) & Y & 8.0 & 512*320 & 2.0s & 1547 & 5.69\% \\
CogVideoX (5B) \citep{yang2024cogvideox} (f) & Y & 10.0 & 720*480 & 4.0s & 1786 & 6.57\% \\
OpenSora-Plan (v1.3) \citep{lin2024open} (u) & Y & 18.0 & 640*352 & 5.2s & 1378 & 5.07\% \\
\midrule
\multicolumn{7}{c}{\textbf{Tier3: Moderate.} 11349 videos, 41.77\%} \\ 
\midrule
CogVideoX (2B) \citep{yang2024cogvideox} (e) & Y & 10.0 & 720*480 & 4.0s & 1774 & 6.53\% \\
LTX-Video-0.9.5 \citep{HaCohen2024LTXVideo} (z) & Y & 25.0 & 704*480 & 4.8s & 1692 & 6.23\% \\
OpenSora (v1.2) \citep{opensora} (x) & Y & 8.0 & 640*480 & 1.6s & 907 & 3.34\% \\
Latte \citep{ma2025lattelatentdiffusiontransformer} (b) & Y & 10.0 & 512*512 & 1.6s & 1510 & 5.56\% \\
VideoCrafter2 \citep{chen2024videocrafter2} (n) & Y & 10.0 & 512*320 & 1.6s & 1172 & 4.31\% \\
Vchitect-2.0 \citep{fan2025vchitect} (p) & Y & 10.0 & 512*320 & 1.6s & 1235 & 4.55\% \\
AnimateDiff \citep{guo2023animatediff} (a) & Y & 10.0 & 512*512 & 2.4s & 1755 & 6.46\% \\
Hotshot-XL \citep{Mullan_Hotshot-XL_2023} (m) & Y & 8.0 & 673*384 & 1.0s & 1304 & 4.80\% \\
\midrule
\multicolumn{7}{c}{\textbf{Tier4: Poor / Early.} 3407 videos, 12.54\%} \\ 
\midrule
ModelScope \citep{wang2023modelscopetexttovideotechnicalreport} (d) & Y & 10.0 & 256*256 & 2.4s & 967 & 3.56\% \\
LTX-Video-0.9.1 \citep{HaCohen2024LTXVideo} (i) & Y & 10.0 & 704*512 & 3.3s & 1333 & 4.91\% \\
ZeroScope \citep{zeroscope} (j) & Y & 10.0 & 256*256 & 2.4s & 395 & 1.45\% \\
T2V-Zero \citep{text2video-zero} (k) & Y & 10.0 & 256*256 & 0.8s & 712 & 2.62\% \\
\midrule
\multicolumn{7}{c}{\textbf{All: 27168 videos}} \\ 
\bottomrule

\end{tabular}
}

\label{tab:apdx_video_stat}
\end{table}

\clearpage
\subsection{Video Examples from Different Quality Tiers.}
\label{subsec:apdx_video_examples}
Below we show some videos of each quality tier, from "Perfect / Modern" to "Poor / Early". 

\begin{figure}[!ht]
    \centering
    \includegraphics[width=0.99\textwidth]{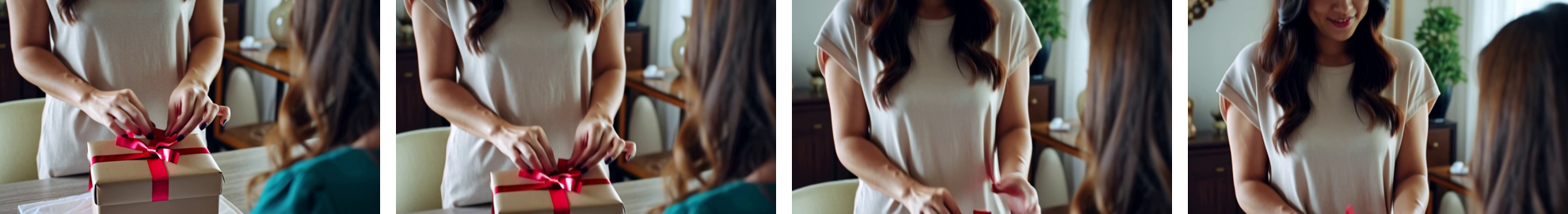}
    \caption*{Example video of the quality tier \textbf{``Perfect / Modern"}, from Kling-1.6. Prompt is: \textit{A woman ties a red ribbon around a gift box, carefully wraps it in shiny paper, and then smiles as she hands it to her friend.}}
    \label{fig:example1}
\end{figure}

\begin{figure}[!ht]
    \centering
    \includegraphics[width=0.99\textwidth]{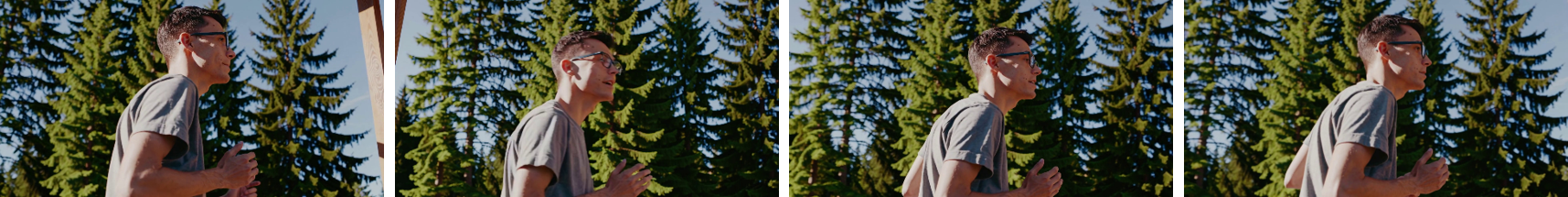}
    \caption*{Example video of the quality tier \textbf{``Perfect / Modern"}, from Sora. Prompt is: \textit{A young man in glasses and a gray t-shirt stands on a wooden deck, gesturing with his hands and expressing different emotions. The background shows a scenic forest with tall trees and a clear sky. His facial expressions change as he moves his hands, sometimes near his face, indicating various reactions. The camera captures him in a steady medium shot, focusing on his upper body and gestures.}}
    \label{fig:example2}
\end{figure}

\begin{figure}[!ht]
    \centering
    \includegraphics[width=0.99\textwidth]{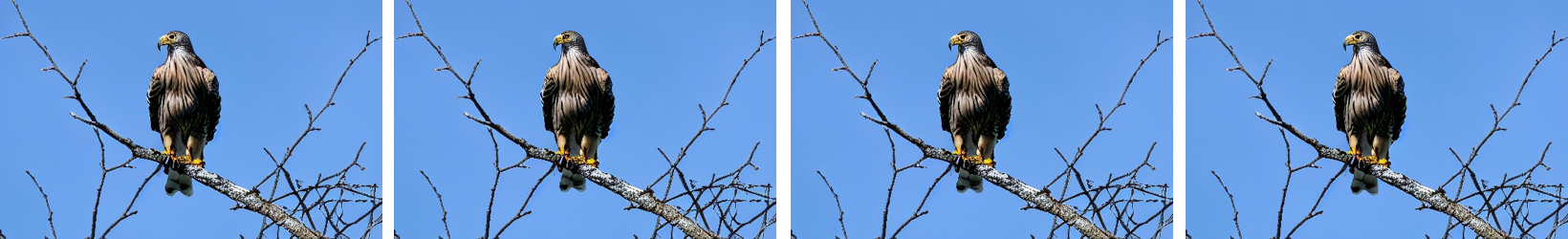}
    \caption*{Example video of the quality tier \textbf{``Good"}, from LaVie-base. Prompt is: \textit{A hawk perches on a leafless tree branch, facing away from the camera and gazing up at the clear blue sky. The calm scene features a few wispy clouds and barren tree branches. The hawk remains still, with occasional head movements, set against a peaceful, natural backdrop.}}
    \label{fig:example3}
\end{figure}

\begin{figure}[!ht]
    \centering
    \includegraphics[width=0.99\textwidth]{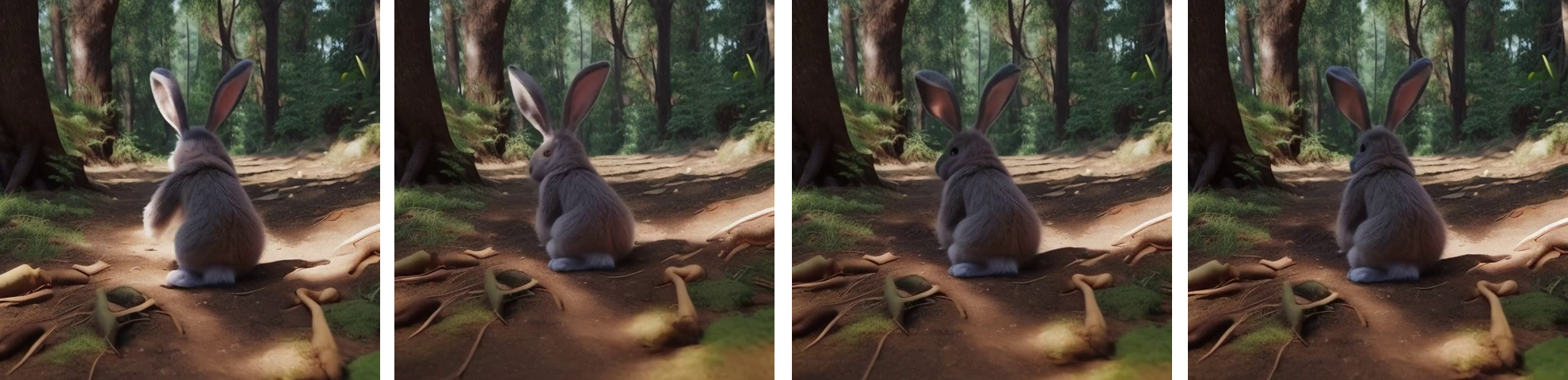}
    \caption*{Example video of the quality tier \textbf{``Good"}, from MagicTime. Prompt is: \textit{A scene showing the lost bunny, its eyes wide with fear, as it navigates through a dense forest, with Sammy guiding it safely home.}}
    \label{fig:example4}
\end{figure}

\begin{figure}[!ht]
    \centering
    \includegraphics[width=0.99\textwidth]{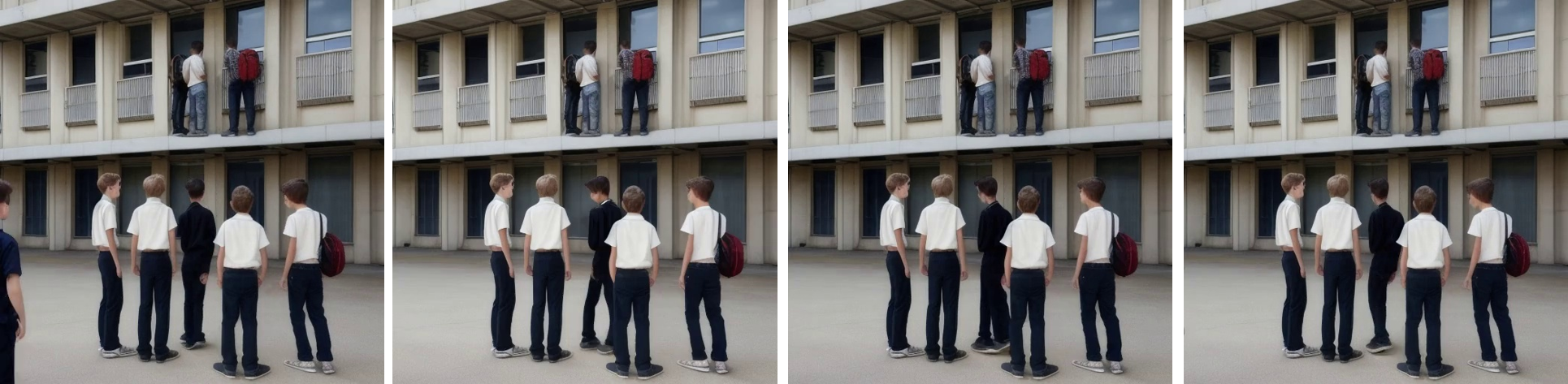}
    \caption*{Example video of the quality tier \textbf{``Moderate"}, from AnimateDiff. Prompt is: \textit{5 boys of age 17 standing outside a school building. Three of them are looking at other students passing by, one is looking at his mobile, and two are talking to each other. Crane up.}}
    \label{fig:example5}
\end{figure}

\begin{figure}[!ht]
    \centering
    \includegraphics[width=0.99\textwidth]{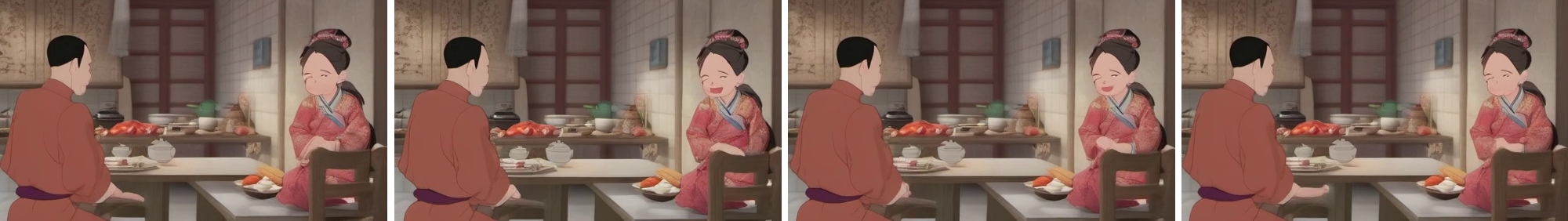}
    \caption*{Example video of the quality tier \textbf{``Moderate"}, from Hotshot-XL. Prompt is: \textit{A cozy family kitchen with breakfast items on the table. Xiao Ming, wearing traditional home attire, engages in a lively conversation with his parents, emphasizing the warmth of family bonds and traditional values.}}
    \label{fig:example6}
\end{figure}

\begin{figure}[!ht]
    \centering
    \includegraphics[width=0.99\textwidth]{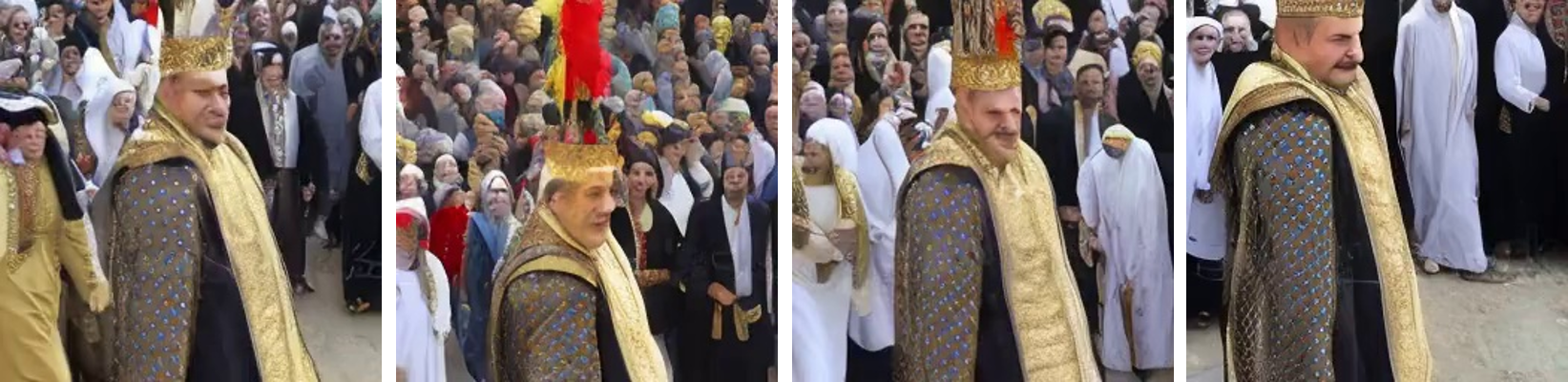}
    \caption*{Example video of the quality tier \textbf{``Poor / Early"}, from ModelScope. Prompt is: \textit{The Kurdish king wears a crown of gold on his head in 1850. He is imposing, serious, authoritative, loving, tall, and handsome. He walks among the people in Kurdish clothes. Tilt down.}}
    \label{fig:example7}
\end{figure}

\begin{figure}[!ht]
    \centering
    \includegraphics[width=0.99\textwidth]{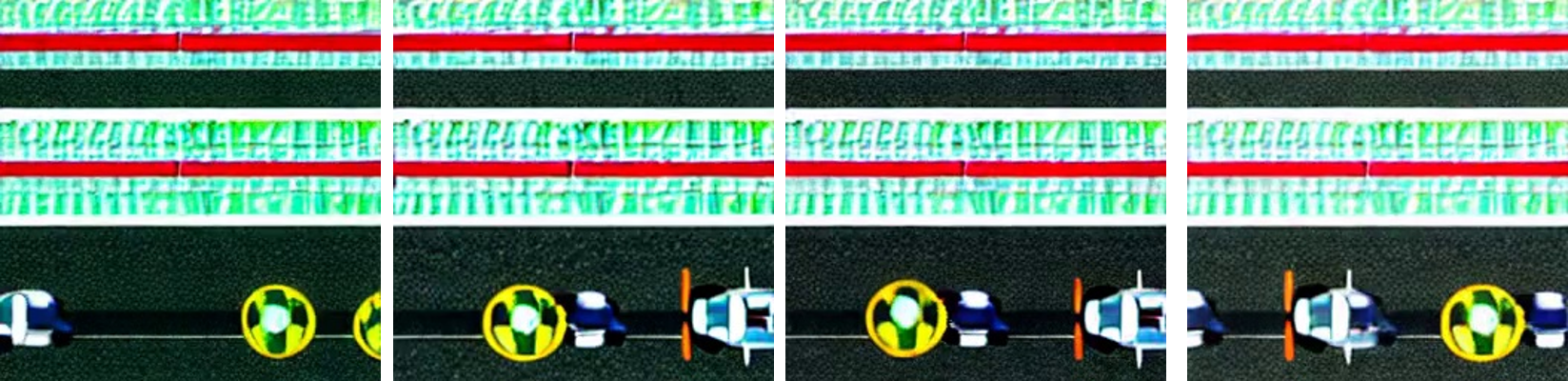}
    \caption*{Example video of the quality tier \textbf{``Poor / Early"}, from Text2Video-Zero. Prompt is: \textit{A busy highway with cars and trucks moving in both directions under a clear blue sky. The scene, filmed from a moving vehicle, highlights a white van with 'Martinez returns from Florida' on its side.}}
    \label{fig:example8}
\end{figure}

\clearpage
\subsection{Annotation Details.}
\label{subsec:apdx_anno_details}
\paragraph{Main Instructions} The main instruction required annotators to assign a score for each dimension based on its definition and to provide a short comment describing the issues observed. For instance, under \textit{Visual Quality}, comments could include ``low resolution,'' ``local blur,'' or ``brightness flicker.'' For \textit{Text-to-Video Alignment}, annotators were asked to note missing elements from the prompt, while for \textit{Physical/Common-Sense Consistency}, they were instructed to highlight any violations of physical laws, common sense, or abnormal artifacts.  

\paragraph{Detailed Guidelines} In addition, we provided detailed annotation guidelines to ensure consistency:  
(i) if a dimension was rated 5, the comment could be omitted, since in such cases a template-based rationale could be generated;  
(ii) if a video was entirely black or unrecognizable, it should be skipped.  
Dimension-specific clarifications were also given:  
\begin{itemize}
    \item \textbf{Visual Quality:} Videos scoring 5 should look almost perfect, comparable to real footage; while videos scoring 1 corresponds to severe flaws, where the subject, object, or motion is hardly identifiable, or strong distortion/disconnection is present.  
    \item \textbf{Text-to-Video Alignment:} For prompts with multiple actions (e.g., “Open the refrigerator, put the elephant in, and close the door”), all actions must be checked for faithful realization. While alignment often correlates with visual quality, clear and smooth videos may still fail to match the prompt. Annotators were instructed to focus on whether the prompt content was expressed correctly, ignoring minor extra details unless they severely misled the meaning.  
    \item \textbf{Physical/Common-Sense Consistency:} Most videos contain at least minor physical issues, but the severity varies. If a prompt itself is unrealistic or absurd, annotators were instructed to disregard this and judge the video independently. Complex reasoning was unnecessary; everyday common sense was considered sufficient for evaluation. 
\end{itemize}

Furthermore, annotators are informed that \textbf{each batch of 10 videos they see sequentially corresponded to the same prompt} but came from different T2V models with diverse quality levels, enabling fairer and more calibrated scoring.

% =======================
\clearpage
\subsection{Prompt Templates for Annotation Processing}
\label{subsec:apdx_prompt_template}

Table~\ref{tab:apdx_prompt_thinking_cmt} shows the prompt template in LLM augmented scoring for eliciting detailed thinking from human annotated quality comments. Table~\ref{tab:apdx_prompt_edit_thinking} shows the prompt template for revising analysis process when the human-annotated score and then adjusted model score are inconsistent with the thinking model’s output analysis. Table~\ref{tab:apdx_prompt_query_sft} shows the prompt template for building query in SFT data and running inference.

\begin{table}[!h]
\renewcommand{\arraystretch}{1.2}
\small
\centering
\caption{Prompt template in LLM augmented scoring for eliciting detailed thinking from human annotated quality comments.}
% \scalebox{0.8}{
\resizebox{\textwidth}{!}{
\renewcommand{\arraystretch}{1.4}
\begin{tabular}{p{1.2\textwidth}}
\toprule

We are collecting and processing human annotations for the quality evaluation of AI-generated videos. \\

\\

\textbf{Dimension definitions:} \\

(1) Visual Quality: \\

Mainly evaluates the video's visual and optical properties, including 'resolution, overall clarity, local blurriness, smoothness, stability of brightness/contrast, distortion/misalignment, abrupt changes, and any other factors the affect the watching experience'. The keywords written by the annotators are also mostly derived from the above factors. \\

(2) Text Alignment:  \\

Mainly assesses whether the generated video fully and accurately depicts the elements mentioned in the text prompt, such as characters, actions, animals, etc., as well as background, quantity, color, weather, and so on. So the keywords written by annotators sometimes only indicate the elements that are missing from the video. \\

(3) Physical/Common-sense Consistency: \\

Mainly examines whether there are any violations of common sense, physical laws, or any other aspects in the video that appear strange or unnatural. Most of the keywords provided by annotators point out the specific abnormalities or inconsistencies they observed in the video. \\

\\

With the reference of some frames of the video, and the comments of 3 dimensions from a human annotator may also be provided, please do your best to analyze and give a INTEGAR score between 1 and 5 for these dimensions, where 1 means very bad, 3 means medium, and 5 means very good. \\

Sometimes human comments may be brief or lacking details, or the human comments may be null, — please check the aspects in dimension definitions and make sure to thoroughly perceive and analyze the video on your own.  
\textbf{Your reasoning should be detailed, professional, and comprehensive. **DO NOT mention any human comment in your thinking**}; you should pretend not to know these comments (if they are provided), they are provided solely to inform and enhance your understanding for better evaluation. \\

\\

\textbf{Output format}: \\

Your response must follow the format below strictly: 

\{

``score\_visual": ``quality score" (this field is only allowed to be a number between 1 and 5, inclusive, ), \\

``score\_t2v": ``quality score" (this field is only allowed to be a number between 1 and 5, inclusive), \\

``score\_phy": ``quality score" (this field is only allowed to be a number between 1 and 5, inclusive), \\

\}

DO NOT include any text before or after the json block. \\

\\

\textbf{Here is the Input:} \\
Text prompt used to generate the video: \$prompt \\

Comment for ``visual quality": \$comment\_visual \\

Comment for ``text-to-video alignment" (the elements or events not expressed or not aligned in the video): \$comment\_t2v  \\

Comment for ``physical/common-sense consistency" (the elements or events that look weird, abnormal or unnatural): \$comment\_phy \\

\bottomrule 

\end{tabular}
}

\label{tab:apdx_prompt_thinking_cmt}
\end{table}

\begin{table}[!h]
\renewcommand{\arraystretch}{1.2}
\small
\centering
\caption{Prompt template for revising analysis process when the human-annotated score and then adjusted model score are inconsistent with the thinking model’s output analysis..}
% \scalebox{0.8}{
\resizebox{\textwidth}{!}{
\setlength{\tabcolsep}{6pt}
\renewcommand{\arraystretch}{1.4}
\begin{tabular}{p{1.2\textwidth}}
\toprule
I'm conducting a multi-dimensional quality assessment of AI-generated videos, focusing on the dimensions of (1) Visual Quality, (2) Text Alignment, and (3) Physical/Common-sense Consistency. \\

\\

I will provide a multi-dimensional quality analysis for a video. However, the scores assigned in the analysis may not be entirely accurate. And the ground truth scores for each dimension will also be provided. Your task is to adjust the analysis text accordingly to ensure it aligns with the actual scores. In many cases, this means revising the severity of issues for certain dimension based on the ground truth scores. The scale of score is [1, 2, 3, 4, 5]. \\

\\

\textbf{**Important Notes:**} \\

(1) \textbf{**Any human comment should NOT be mentioned in the output analysis**}. If the input analysis quote or mention human comments, you should pretend not to know them in your output, they are provided solely to inform and enhance your understanding for better evaluation. \\
 
(2) \textbf{**DO NOT** alter the overall structure or core meaning of the analysis}. Only revise specific expressions or phrases as needed so that the content reasonably reflects the provided scores.  \\

(3) The input original analysis is constructed from the sampled frames of the video, if the input analysis includes evaluations of individual frames or frame-by-frame assessments, you should appropriately transform them into an overall evaluation of the entire video, since the final output is expected to be based on the video as a whole.  \\

(4) Your output analysis should be approximately the same length as the input analysis. If the input analysis is not very detailed and specific, you may extend your output accordingly. \\

\\ 

\textbf{Output format:} \\

Your response must follow the format below strictly: \\
\{ 
    ``new\_thinking": ``modified analysis" (this field is only allowed to be string),
\} 
\\
DO NOT include any text before or after the dictionary block. \\

\\

\textbf{Here is the input:} \\
multi-dimensional analysis: \$thinking

ground truth score of Dim-1 ``Visual Quality":\$v\_score \\

ground-truth scoreof Dim-2 ``Text-to-Video Alignment":\$t\_score \\
 
ground-truth of Dim-3 ``Physical Consistency' (also referred to as Common-sense Consistency): \$p\_score \\

\bottomrule 

\end{tabular}
}

\label{tab:apdx_prompt_edit_thinking}
\end{table}

\begin{table}[!h]
\renewcommand{\arraystretch}{1.2}
\small
\centering
\caption{Prompt template for building query in SFT data and running inference.}
% \scalebox{0.8}{
\resizebox{\textwidth}{!}{
\renewcommand{\arraystretch}{1.4}
\begin{tabular}{p{1.2\textwidth}}
\toprule
We would like to evaluate its quality from three dimensions: 'visual quality', 'text-to-video alignment' and 'physical/common-sense consistency'. 
\\
Below is the definition of each dimension: \\

(1) visual quality: \\ 

The dimension 'visual quality' cares about the video's visual and optical properties, including 'resolution, overall clarity, local blurriness, smoothness, stability of brightness/contrast, distortion/misalignment, abrupt changes, and any other factors the affect the watching experience'. The keywords written by the annotators are also mostly derived from the above factors. \\

(2) text alignment:\\

The dimension 'text-to-video alignment' mainly assesses whether the generated video fully and accurately depicts the elements mentioned in the text prompt, such as characters, actions, animals, etc., as well as background, quantity, color, weather, and so on. So the keywords written by annotators sometimes only indicate the elements that are missing from the video. \\

(3) physical/common-sense consistency:\\

The dimension 'physical/common-sense consistency' mainly examines whether there are any violations of common sense, physical laws, or any other aspects in the video that appear strange or unnatural. Most of the keywords provided by annotators point out the specific abnormalities or inconsistencies they observed in the video.\\

\\

Here we provide an AI video generated by text-to-video models and its text prompt: \\

\$t2v\_prompt. \\

Based on the video content and the dimension definitions, please evaluate the video quality and give the quality score. The score must be in the range of 1 - 5.\\

\bottomrule 

\end{tabular}
}

\label{tab:apdx_prompt_query_sft}
\end{table}
\clearpage
\section{Evaluation Suite}

% =======================
\subsection{Dimension Matching and Modification in Out-of-Domain Benchmarks}
\label{subsec:apdx_modify_ood_bench}

Since different benchmarks define varying dimensions and scoring scales, we align them with the three evaluation dimensions of \score\ (visual quality, text alignment, and physical consistency) and, where necessary, rescale their ground-truth scores. Below we summarize the mapping rules for each benchmark.

\paragraph{VideoGenReward Bench.}
This is a pairwise preference benchmark containing 4,691 videos, forming 25,234 video pairs. It evaluates three dimensions---visual quality (VQ), text alignment (TA), and motion quality (MQ)---and also provides an \textit{Overall} preference label indicating which video is better overall. Among these, VQ and TA correspond closely to \score’s first two dimensions (despite slight definitional differences), so these two dimensions are used for this benchmark. For the \textit{Overall} preference, we use the mean of all available dimension scores from \score\ or the baseline method (if the baseline only has one quality score output, then that score is used directly).

\paragraph{T2VQA-DB.}
Originally a human-annotated video quality dataset with 10,000 videos, each labeled with a single quality score in the range [1,100]. We sample 2,000 videos and construct 1,822 pairs by comparing human-annotated scores. Since the dataset provides only one dimension (the final score), we predict preference by averaging all dimension scores from \score or the baseline method (if the baseline only has one quality score output, then that score is used directly).

\paragraph{MJ-Bench-Video.}
This benchmark contains 2,170 videos and adopts a point-score format with five dimensions: \textit{fineness}, \textit{alignment}, \textit{consistency \& coherence}, \textit{safety}, and \textit{bias \& fairness}. We select the first three, which correspond to \score’s three evaluation dimensions. For baselines with only one final score, we “broadcast” this score across multiple dimensions. The benchmark uses a \{0,1,2\} scale, whereas \score\ and other baselines output (or are normalized to) integer scores in [1,5]. Thus, we apply the following mapping, where $x$ denotes the original score of each dimension, $v$, $t$, $p$ denote the rescaled score of for ``visual quality", ``text alignment" and ``physical consistency", respectively:
\[
v = \begin{cases}
0 & \text{if } x \in \{1,2\}, \\
1 & \text{if } x \in \{3,4\}, \\
2 & \text{if } x = 5,
\end{cases} 
\quad
t = \begin{cases}
0 & \text{if } x=1, \\
1 & \text{if } x \in \{2,3\}, \\
2 & \text{if } x \in \{4,5\},
\end{cases}
\quad
p = \begin{cases}
0 & \text{if } x=1, \\
1 & \text{if } x \in \{2,3\}, \\
2 & \text{if } x \in \{4,5\}.
\end{cases}
\]
The benchmark also provides an \textit{Overall} score, for which we again take the mean of all available dimension scores (or the single dimension if only one is provided), rescaled into \{0,1,2\} using the same rule.

\paragraph{VideoPhy2-test.}
This benchmark contains 3,396 videos with two dimensions, SA: \textit{semantic adherence} and PC:  \textit{physical consistency}. These map perfectly to \score’s second and third dimensions. For baselines lacking one of the dimensions (e.g., VideoReward, which provides VQ, TA, MQ but no physical consistency), we skip the missing dimension. The scoring scale is \{1,2,3,4,5\}, so no rescaling is required.

% \paragraph{AIGVE-Bench.}
% This benchmark contains 2,429 videos with multiple fine-grained aspects. To align with \score’s three evaluation dimensions, we aggregate them into equivalent categories: we take the average of \{\textit{technical quality}, \textit{element quality}, and \textit{action quality}\} as the ground-truth label for the \textit{visual quality} dimension; the average of \{\textit{element presence} and \textit{action presence}\} as the label for \textit{text alignment}; and \textit{physics} as the label for \textit{physical consistency}. The original scoring scale is \{1,2,3,4,5\}, so no further rescaling is required.

% =======================
\clearpage
\subsection{Dimension Matching and Score Rescaling for Baselines}
\label{subsec:apdx_baseline_rescaling}

Since different baseline models adopt varying evaluation dimensions and scoring scales, we apply \textbf{dimension matching} and \textbf{score rescaling} to make them compatible with \bench\ and \score. Our goal is to ensure that all baselines output scores on the three dimensions---visual quality (v), text alignment (t), and physical consistency (p)---within a unified range of integers 1-5. A summary of the mapping rules is provided in Table~\ref{tab:apdx_baseline_rescaling}.  

\paragraph{Baseline Dimension Matching.} 
With $v$, $t$, and $p$ to denote the score of visual quality, text alignment, physical/common-sense consistency, respectively, we consider three cases:
\begin{itemize}

    \item \textbf{Broadcast.} Some baselines only output a single final score. In this case, we broadcast the same score to our three dimensions $v$, $t$, and $p$.
    
    \item \textbf{Good Match.} Some baselines already report dimensions that closely match ours, so we directly use their outputs without modification.  
    
    \item \textbf{Customized.} For baselines with different or partially overlapping dimensions, we design customized mappings. 
    \begin{itemize}
        \item \textit{VideoReward}: outputs Visual Quality, Text Alignment, and Motion Quality. We use the outputs of first two dimensions as $v$ and $t$, and skip Motion Quality.  
        \item \textit{AIGVE-MACS}: outputs multiple fine-grained dimensions. We average \{technical quality, element quality, action quality\} as $v$, average \{element presence, action presence\} as $t$, and use physics as $p$. 
        \item \textit{VideoPhy2-Auto-Eval}: outputs Semantic Adherence (SA) and Physical Consistency (PC). We use SA as $t$ and PC as $p$, while skipping $v$.  
    \end{itemize}
\end{itemize}

\paragraph{Baseline Score Rescaling.}  
To make results comparable, we rescale all baseline outputs into a unified integer range of 1--5. A summary of the mapping rules is provided in Table~\ref{tab:apdx_baseline_rescaling}.    

\begin{itemize}
\item Linear Scaling or No Scaling. For baselines with well-defined score ranges (e.g., [0,1], [0,100]), we apply linear normalization followed by rounding to the nearest integer in $\{1,2,3,4,5\}$.  

\item Ordinal categories using Gaussian-distribution quantile thresholds. 
For baselines without fixed score bounds, we adopt an ordinal mapping based on Gaussian-distribution quantile thresholds. Specifically, raw scores are assumed to approximately follow a Gaussian distribution and are divided into five categories using the 20\%, 40\%, 60\%, and 80\% quantiles of the standard normal distribution.  
  If the raw scores typically fall within [-2.0, 2.0] and we assume a Gaussian Distribution $N(0, 1)$, thus apply the following mapping:  
\[
\text{score} =
\begin{cases}
1 & \text{if } z < \Phi^{-1}(0.2), \\
2 & \text{if } \Phi^{-1}(0.2) \leq z < \Phi^{-1}(0.4), \\
3 & \text{if } \Phi^{-1}(0.4) \leq z < \Phi^{-1}(0.6), \\
4 & \text{if } \Phi^{-1}(0.6) \leq z < \Phi^{-1}(0.8), \\
5 & \text{otherwise},
\end{cases}
\]
where $z$ is the raw model score and $\Phi^{-1}$ denotes the inverse CDF of the standard Gaussian.  
    \begin{itemize}
    \item \textit{ImageReward} and \textit{VisionReward}: most scores are in [-2.0, 2.0], assume $N(0, 1)$ and follow the mapping above. 
    \item \textit{VideoReward}: most scores are in [-3.0, 3.0], so we assume a Gaussian Distribution $N(0, 1.5)$, and $z$ is replaced by $z/1.5$ in the rules above to firstly normalize the raw score before converting it to integers.
    \end{itemize}

\end{itemize}

\begin{table}[h]
\small
\centering
\caption{Rescale output scores and map dimensions of baselines models to align with our \score and \bench.}
\scalebox{0.85}{
% \resizebox{\textwidth}{!}{
\setlength{\tabcolsep}{5pt}
\renewcommand{\arraystretch}{1.5}
\begin{tabular}{c|c|c|l}
\toprule
     Model   & Dimension Mapping & Original Scale       & Score Rescaling Method   \\
\midrule
\multicolumn{4}{c}{\textbf{Reward/Scoring Models for Image} (averaged on sampled frames)} \\ 
\midrule
ImageReward     & Broadcast      & most in {[}-2.0,2.0{]} &  Ordinal categories using Gaussian-distribution quantile thresholds. \\
DeQA-Score            & Broadcast      & {[}0.0, 5.0{]}                         & Linearly amplify and round                                                     \\
Q-Insight       & Good Match      & {[}1.0, 5.0{]}                         & Linearly amplify and round   \\
\midrule
\multicolumn{4}{c}{\textbf{Reward/Scoring Models for Video}} \\ 
\midrule
VideoReward     & Customized      & most in {[}-4.0,4.0{]} &  Ordinal categories using Gaussian-distribution quantile thresholds. \\
UnifiedReward   & Good Match     & \{1,2,3,4,5\}                          & No rescaling                                                                   \\
VisionReward    & Broadcast      & most in {[}-1.0,1.0{]} & Ordinal categories using Gaussian-distribution quantile thresholds. \\
Q-Align         & Broadcast      & {[}0.0, 1.0{]}                         & Linearly amplify and round                                                     \\
AIGVE-MACS      & Customized        & \{1,2,3,4,5\}                          & No rescaling                                                                   \\
VideoPhy2      & Customized        & \{1,2,3,4,5\}                          & No rescaling                                                                   \\
Dover           & Broadcast      & {[}0.0, 1.0{]}                         & Linearly amplify and round                                                     \\

\bottomrule
\end{tabular}
}

\label{tab:apdx_baseline_rescaling}
\end{table}
\clearpage
\section{Full Evaluation Results}
\label{sec:apdx_full_res}

% =======================
\subsection{Full Results on VideoGen-Reward-Bench}
\label{subsec:apdx_full_res_vgrbench}
VideoGen-Reward-Bench is a video preferenc over three dimensions: visual quality, text alignment, and motion quality. The task is to compare a pair of videos and judge which one is better along these axes. Among them, the first two dimensions are broadly aligned with ours, while the benchmark also provides an additional measure of overall preference. 

For the preference benchmarks, we report results under two settings. The w/ ties version includes all test entries, where in some cases the two compared videos (including the ground-truth reference) are judged as equally preferred. The w/o ties version is a subset obtained by removing those entries with equal preference labels. The full evaluation results of preference prediction accuracy are shown in Table~\ref{tab:apdx_full_res_vgrbench}. 

\begin{table}[h]

\small
\centering
\caption{Full evaluation results on \textbf{VideoGen-Reward-Bench}. \textbf{Bold} denotes the best model and the \underline{underlined} denotes the second best.}
\scalebox{0.95}{
% \resizebox{\textwidth}{!}{
\renewcommand{\arraystretch}{1.5}
\setlength{\tabcolsep}{4pt}

\begin{tabular}{c|cc|cc|cc}
\toprule
  \multirow{2}{*}{\textbf{VideoGen-Reward-Bench}}  & \multicolumn{2}{c|}{Visual Quality} & \multicolumn{2}{c|}{Text Alignment} & \multicolumn{2}{c}{Overall} \\
    \cmidrule(lr){2-3} \cmidrule(lr){4-5} \cmidrule(lr){6-7}           
& w ties          & w/o ties         & w ties          & w/o ties         & w ties      & w/o ties      \\
\midrule
\multicolumn{7}{c}{\textbf{Reward/Scoring Models for Image} (averaged on sampled frames)} \\ \midrule
ImageReward        & 31.64           & 51.40            & 44.00           & 60.72            & 47.14       & 58.61         \\
DeQA-Score              & 41.07           & \underline{69.55}            & 36.22           & 53.23            & 53.88       & 67.91         \\
Q-Insight          & 30.68           & 66.34            & 42.11           & 59.47            & 54.05       & 66.34   \\
\midrule
\multicolumn{7}{c}{\textbf{Reward/Scoring Models for Video}} \\ \midrule
VideoScore-v1.1    & \underline{47.41}           & 30.84            & 26.09           & 30.85            & 16.79       & 40.19         \\
VideoReward        & \textbf{53.21}           & \textbf{75.58}            & \textbf{52.75}           & \textbf{72.18}           & \textbf{59.69}       & \textbf{73.66}         \\
UnifiedReward      & 41.27           & 39.42            & 40.11           & 36.58            & 53.31       & 58.83         \\
VisionReward       & 35.89           & 59.03            & 44.86           & 61.15            & 54.31       & 67.58         \\
Q-Align            & 32.01           & 52.98            & 35.77           & 51.06            & 42.05       & 52.52         \\
AIGVE-MACS         & 38.05           & 30.80            & 30.76           & 11.66            & 37.09       & 37.08         \\
VideoPhy2 & -               & -                & 37.04           & 22.14            & 30.75       & 26.41         \\
Dover              & 39.34           & 68.87            & 38.01           & 55.65            & 54.27       & \underline{68.58}         \\
\midrule
\multicolumn{7}{c}{\textbf{Ours}} \\ 
\midrule
\score (SFT only) & 37.74 & 63.17 & 43.07 & 61.35 & 50.79 & 63.80 \\
\score (RL w/o SFT) & 34.67 & 65.87 & \underline{48.70} & \underline{65.92} & \underline{54.53} & 65.59 \\
\score (SFT $+$ RL) & 37.44 & 63.08 & 42.87 & 60.61 & 51.53 & 63.72 \\
    \bottomrule
\end{tabular}
}

\label{tab:apdx_full_res_vgrbench}
\end{table}

% =======================
\clearpage
\subsection{Full Results on MJ-Bench-Video}
\label{subsec:apdx_full_res_mj}
To maximize compatibility with the evaluation dimensions of \score, we selected three aspects from MJ-Bench-Video that are most semantically aligned: Fineness, Alignment, and Coherence \& Consistency. These aspects correspond respectively to the three dimensions in \score: visual quality, text alignment, and physical/commonsense consistency.

The full evaluation results of the three aspects and the overall scores are shown in Table~\ref{tab:apdx_full_res_mj}, with prediction accuracy between model outputs and ground truths adopted as metrics.

\begin{table}[h]

\small
\centering
\caption{Full evaluation results on \textbf{MJ-Bench-Video}. \textbf{Bold} denotes the best model and the \underline{underlined} denotes the second best.}
\scalebox{0.85}{
\renewcommand{\arraystretch}{1.5}
\setlength{\tabcolsep}{6pt}
\begin{tabular}{l|ccc|c}
\toprule
\multirow{3}{*}{\textbf{MJ-Bench-Video}} & \multicolumn{4}{c}{Accuracy}  \\  
\cmidrule{2-5}
      & \multirow{2}{*}{Fineness}  & \multirow{2}{*}{Alignment}  & Coherence \& & \multirow{2}{*}{Overall}   \\
 & & & Consistency & \\
  \midrule
\multicolumn{5}{c}{\textbf{Reward/Scoring Models for Image} (averaged on sampled frames)} \\ \midrule
ImageReward & 47.05 & 28.07 & 29.03 & 37.51 \\
DeQA-Score   & 18.57 & 51.20 & 52.40 & 44.19 \\
Q-Insight  & 12.72 & 42.86 & 28.07 & 52.58 \\
\midrule
\multicolumn{5}{c}{\textbf{Reward/Scoring Models for Video}} \\ 
\midrule
VideoScore-v1.1 & 13.69 & \textbf{64.19} & \textbf{79.22} & \textbf{71.57} \\
VideoReward  & \textbf{79.36} & 38.99 & -     & 51.75 \\
UnifiedReward  & 43.50 & 21.98 & 18.16 & 23.18 \\
VisionReward   & 36.31 & \underline{55.99} & \underline{67.51} & 56.91 \\
Q-Align   & 14.77 & 31.74 & 26.41 & 21.97 \\
AIGVE-MACS  & 20.18 & 26.27 & 21.39 & 31.00 \\
VideoPhy2-Auto-Eval  & -     & 38.97 & 7.89  & 24.00 \\
Dover  & 29.26 & 45.67 & 48.02 & 43.69 \\
\midrule
\multicolumn{5}{c}{\textbf{Ours}} \\ 
\midrule
\score (SFT only) & 33.95 & 46.20 & 57.80 & \underline{66.88} \\
\score (RL w/o SFT) & \underline{64.68} & 32.79 & 57.27 & 56.43 \\
\score (SFT + RL) & 22.50 & 48.58 & 66.79 & 65.77 \\

\bottomrule
\end{tabular}
}

\label{tab:apdx_full_res_mj}
\end{table}

% =======================
\clearpage
\subsection{Full Results on VideoPhy2-test}
\label{subsec:apdx_full_res_videophy2}
Video-Phy2-Test is a human-annotated test set with two dimensions: semantic adherence and physical consistency (abbreviated as semantic and physical in our tables). These two dimensions correspond directly to the latter two evaluation dimensions in our framework.

The full evaluation results of the two dimensions are shown in Table~\ref{tab:apdx_full_res_videophy2}, with prediction accuracy and PLCC between model outputs and ground truths adopted as metrics.

\begin{table}[h]

\small
\centering
\caption{Full evaluation results on \textbf{Video-Phy2-test}. \textbf{Bold} denotes the best model and the \underline{underlined} denotes the second best.}
\scalebox{0.95}{
\renewcommand{\arraystretch}{1.5}
\setlength{\tabcolsep}{4pt}

\begin{tabular}{l|ccc|ccc}
\toprule
    \multirow{2}{*}{\textbf{Video-Phy2-test}}               & \multicolumn{3}{c}{Accuracy} & \multicolumn{3}{c}{PLCC}        \\
                   & Semantic  & Physical & Avg   & Semantic & Physical & Avg     \\
\midrule
\multicolumn{7}{c}{\textbf{Reward/Scoring Models for Image} (averaged on sampled frames)} \\ \midrule
ImageReward        & 23.73     & 19.23    & 21.48   & 15.28    & 3.07     & 9.18  \\
DeQA-Score               & 28.74     & 28.96    & 28.85  & 3.55     & 2.14     & 2.85  \\
Q-Insight          & 29.21     & \underline{32.59}    & 30.90  & 22.45    & 4.98     & 13.72 \\
\midrule
\multicolumn{7}{c}{\textbf{Reward/Scoring Models for Video}} \\ 
\midrule
VideoScore-v1.1    & 29.81     & 26.08    & 27.95  & 11.61    & 13.09    & 12.35 \\
VideoReward        & 31.33     & -        & 31.33 & 34.54    & -        & \textbf{34.54} \\
UnifiedReward      & 17.64     & 26.39    & 22.02 & 34.57    & \underline{22.78}    & 28.68 \\
VisionReward       & 31.95     & 13.20    & 22.58  & 28.11    & 13.67    & 20.89 \\
Q-Align            & 18.43     & 28.00    & 23.22   & 5.52     & 2.70     & 4.11  \\
AIGVE-MACS         & 12.23     & 21.63    & 16.93   & 8.09    & 11.90     & 10.00  \\
VideoPhy2-Auto-Eval & \textbf{37.96}     & \textbf{37.31}    & \textbf{37.64} & 38.64    & \textbf{29.84}    & \underline{34.24} \\
Dover              & 26.56     & 29.86    & 28.21  & 3.85     & 1.15     & 2.50  \\
\midrule
\multicolumn{7}{c}{\textbf{Ours}} \\ 
\midrule
\score (SFT only) & 32.24 & 27.80 & 30.02 & 27.22 & 13.85 & 20.54 \\
\score (RL only) & 31.71 & 23.66 & 27.69 & \underline{39.07} & 16.90 & 27.99 \\
\score (SFT + RL) & \underline{37.48} & 29.67 & \underline{33.58} & \textbf{41.08} & 17.57 & 29.33\\

\bottomrule
\end{tabular}
}

\label{tab:apdx_full_res_videophy2}
\end{table}

% % =======================
\clearpage
% \subsection{Full Results on AIGVE-Bench}
% \label{subsec:apdx_full_res_aigve}
% We choose visual quality, text alignment and physical consistency in AIGVE-Bench, which match perfectly with \score.
% The full evaluation results of the three dimensions are shown in Table~\ref{tab:apdx_full_res_videophy2}, with prediction accuracy, relaxed accuracy (counting cases where the prediction differs from the ground truth by at most one point) and PLCC between model outputs and ground truths adopted as metrics.
%  \input{tables/apdx_full_res_aigve}

% =======================
\clearpage
\subsection{Full Results of Best-of-N sampling on VBench}
\label{subsec:apdx_full_res_BoN}

\begin{table*}[!h]
\small
\centering
\caption{Quality evaluatioin of eight T2V models on V-Bench \textbf{with} BoN sampling by our \score, compared with random ones. We can see consistent improvement. }
\renewcommand{\arraystretch}{1.5}
\resizebox{\textwidth}{!}{
\setlength{\tabcolsep}{4pt}
\begin{tabular}{l|cc|cccccccccc}
\toprule
\multirow{3}{*}{\textbf{\large Best-of-N}}  &  \multicolumn{10}{c}{Dimensions in \textbf{VBench}}  \\ 
\cline{2-13}
 & \multicolumn{2}{c|}{\textbf{Average}} & \multicolumn{2}{c}{Subject } & \multicolumn{2}{c}{Background} & \multicolumn{2}{c}{Aesthetic} & \multicolumn{2}{c}{Imaging} & \multicolumn{2}{c}{Motion} \\

 \cmidrule(lr){2-3}  \cmidrule(lr){4-5}   \cmidrule(lr){6-7}  \cmidrule(lr){8-9}  \cmidrule(lr){10-11}  \cmidrule(lr){12-13}    
 
       & Random & \textbf{BoN} &  Random & \textbf{BoN} & Random & \textbf{BoN} & Random & \textbf{BoN} & Random & \textbf{BoN} & Random & \textbf{BoN}\\ 
\midrule

% VideoCrafter2 & 83.83 & \textbf{83.93} & 96.84          & \textbf{97.10} & 97.55 & \textbf{97.68} & 59.25          & \textbf{59.66} & \textbf{67.98} & 67.64          & 97.53          & \textbf{97.59} \\

Lavie-base    & 82.85 & \textbf{83.07} & 95.40          & \textbf{95.69} & 96.89 & \textbf{97.08} & 56.64          & \textbf{57.15} & \textbf{67.99} & 67.98          & 97.34          & \textbf{97.47} \\

% HotShot-XL    & 82.77 & \textbf{82.81} & \textbf{95.62} & 95.80          & 95.96 & \textbf{96.10} & \textbf{56.77} & 56.51          & 67.92          & \textbf{68.06} & \textbf{97.59} & 97.56          \\

AnimateDiff   & 81.97 & \textbf{83.15} & 91.16          & \textbf{94.18} & 94.30 & \textbf{95.64} & \textbf{60.90} & 60.28          & \textbf{69.36} & 69.01          & 94.14          & \textbf{96.64} \\

VideoCrafter1 & 80.03 & \textbf{80.63} & 95.35          & \textbf{95.58} & 95.76 & \textbf{96.05} & 46.00          & \textbf{47.67} & 67.03          & \textbf{67.58} & 95.99          & \textbf{96.26} \\

ModelScope    & 78.75 & \textbf{79.70 }            & 93.68          & \textbf{95.07}              & 95.52 & \textbf{96.40 }             & 46.23          & \textbf{47.60  }            & 61.64          & \textbf{62.32}              & 96.66          & \textbf{97.10 }             \\

ZeroScope      & 76.36 & \textbf{77.84}              & 91.32          & \textbf{93.04}              & 94.50 & \textbf{95.37 }             & 45.27          & \textbf{47.55}              & 55.25          & \textbf{56.95}              & 95.48          & \textbf{96.30}              \\

LVDM          & 75.33 & \textbf{76.26} & 88.79          & \textbf{89.91} & 93.14 & \textbf{93.81} & 41.01          & \textbf{42.00} & 60.94          & \textbf{62.24} & 92.75          & \textbf{93.35} \\

\bottomrule
\end{tabular}
}
\vspace{-4pt}
\label{tab:4_best_of_n}
\end{table*}
\clearpage
\section{Experiment Setuo and Ablation Studies}
\label{sec:apdx_exp_and_ablation}

% =======================
\subsection{SFT experiment setup}
\label{subsec:apdx_sft_setup}
We conduct SFT with sampling fps of 2, a maximum frame resolution of 960$\times$720, learning rates of 5e-5, and epochs of 2 with one epoch taking about 6 hours on 8$\times$A800 GPUs.

% =======================
\subsection{Ablation on sampling fps in SFT training}
\label{subsec:apdx_ablation_sample_fps}
During training, videos are sampled at 2 fps, which we find sufficient for evaluation: ``visual quality" primarily reflects global perceptual properties, ``text alignment" focuses on semantic adherence, and most issues of ``physical consistency'' or abnormal events typically last longer than half a second, ensuring they can still be captured at this frame rate. 

We also conduct an ablation on a 17k subset to study the effect of training sampling fps, comparing 2, 4, and 8 fps settings. As shown in Table~\ref{tab:apdx_ablation_train_fps}, increasing the sampling rate does not yield significant performance gains, while it noticeably increases computational cost and training time. Therefore, we adopt 2 fps as the default setting in our main SFT experiments and in all subsequent ablations of other hyper-parameters.

\begin{table}[h]

\small
\centering
\caption{Ablation results on a 17k subset of \score data for different sampling fps in SFT.}
\label{tab:apdx_ablation_train_fps}
\scalebox{0.8}{
\setlength{\tabcolsep}{6pt}
\renewcommand{\arraystretch}{1.5}
\begin{tabular}{c|ccc|c|ccc|c|ccc|c}
\toprule

Train Sampling fps & \multicolumn{4}{c|}{Accuracy} & \multicolumn{4}{c|}{Relaxed Accuracy} & \multicolumn{4}{c}{PLCC}  \\  \cmidrule{2-13}
 (17k subset)                            & Visual  & Align  & Phy  & Avg  & Visual    & Align    & Phy    & Avg    & Visual & Align & Phy & Avg  \\
\midrule
 2fps & 54.67 & 39.33 & 46.67 & 46.89 & 94.00 & 81.33 & 90.00 & 88.44 & 73.62 & 60.24 & 54.72 & 62.86 \\
 4fps & 48.67 & 42.67 & 49.33 & 46.89 & 94.00 & 82.00 & 92.67 & 89.56 & 67.87 & 61.85 & 63.86 & 64.53 \\
 8fps & 51.00 & 45.33 & 48.00 & 48.11 & 92.00 & 85.33 & 88.67 & 88.67 & 64.34 & 65.71 & 52.05 & 60.70  \\
\bottomrule
\end{tabular}
}

\end{table}

% =======================
\clearpage
\subsection{Ablation on learning rate and epochs in SFT training}
\label{subsec:apdx_ablation_lr_epoch}

We perform ablations on two key hyper-parameters: learning rate 1e-5, 2e-5, 5e-5, 1e-4, 2e-4 and epochs $\{1, 2, 3\}$. The results on \bench\ are summarized in Table~\ref{tab:apdx_ablation_lr_epoch}. 

\begin{figure}[htbp]
  \centering
  \begin{minipage}[b]{0.47\textwidth}
    \centering
    \includegraphics[width=\textwidth]{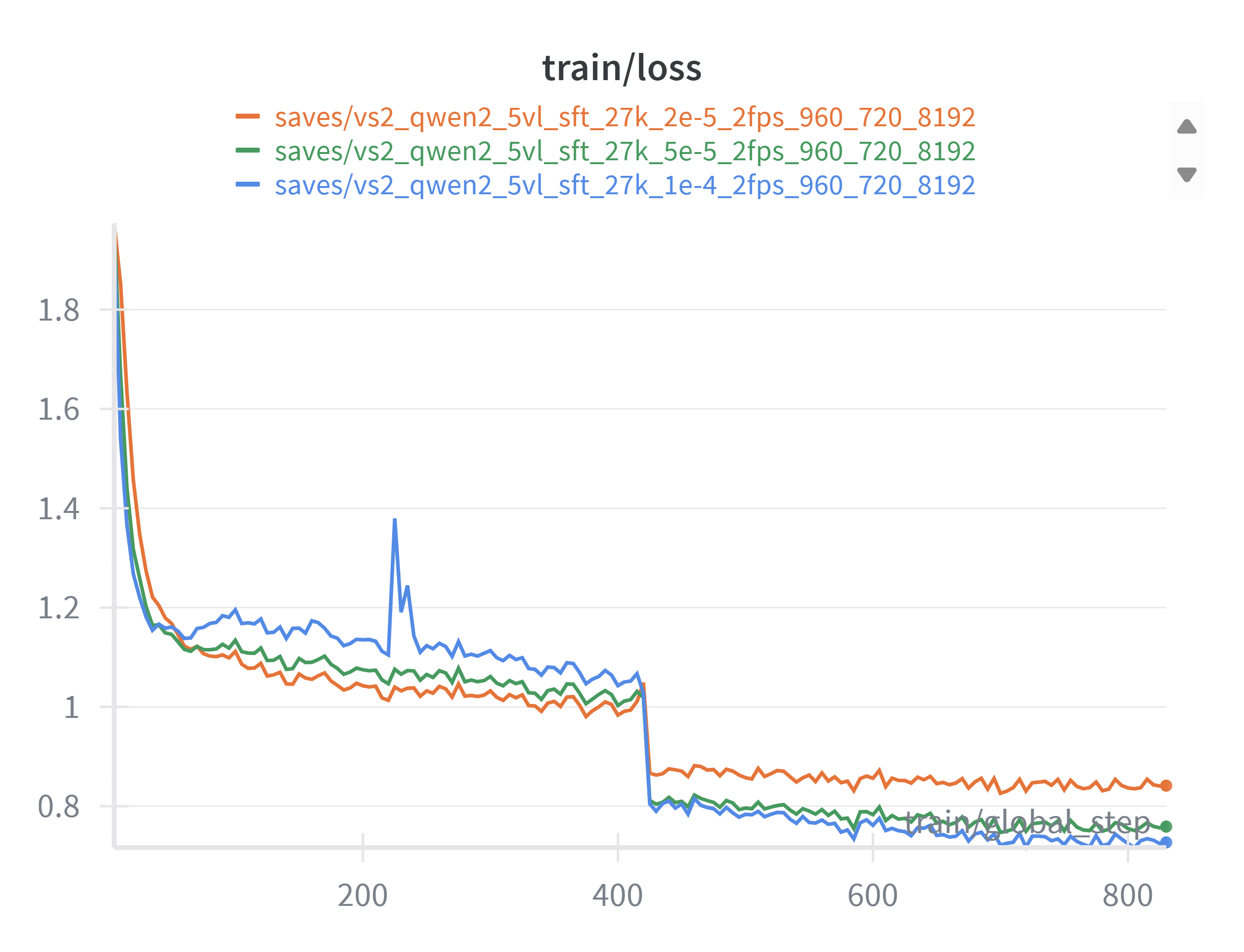}
    \caption{Training loss in ablations of learning rate, 2e-5, 5e-5, and 1e-4 are shown. }
    \label{fig:apdx_loss_lr}
  \end{minipage}
  \hfill
  \begin{minipage}[b]{0.47\textwidth}
    \centering
    \includegraphics[width=\textwidth]{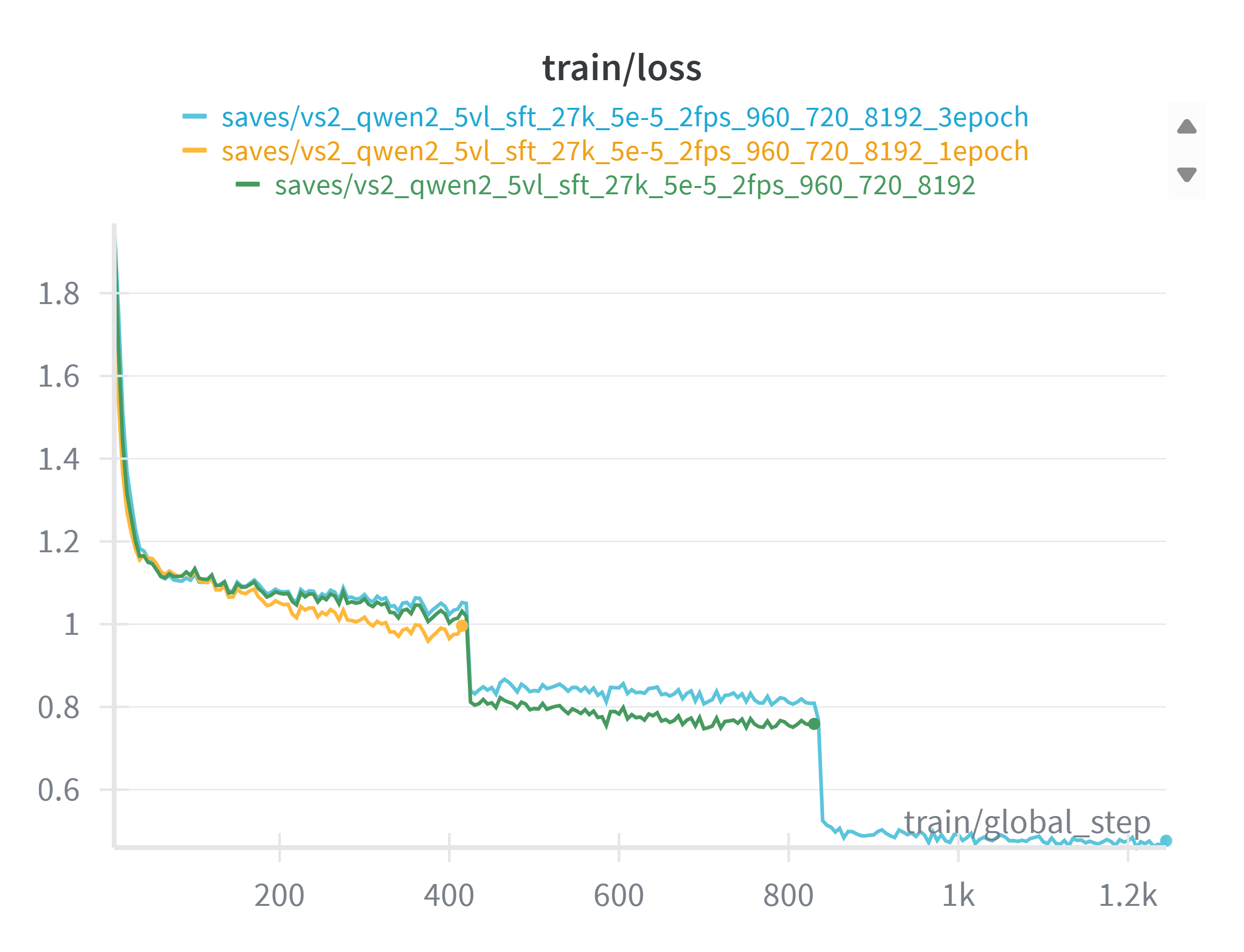}
    \caption{Training loss in ablations of training epoch, 1epoch, 2epoch, and 3epoch are shown.}
    \label{fig:apdx_loss_epoch}
  \end{minipage}
\end{figure}

For learning rate, $1 \times 10^{-4}$ achieves slightly higher accuracy than $5 \times 10^{-5}$, but its loss curve is less stable and shows lower values in the second epoch, as shown in Figure~\ref{fig:apdx_loss_lr} and ~\ref{fig:apdx_loss_epoch}, suggesting potential overfitting, which could harm performance on out-of-domain benchmarks. By contrast, 2e-5 exhibits a much higher loss curve in later stages, indicating underfitting. Balancing in-domain accuracy and loss smoothness, we choose 5e-5 as the default learning rate. 

For epochs, the 2-epoch setting outperforms both 1 and 3 epochs, and is therefore adopted as the main version. This chosen SFT checkpoint also serves as the base model for subsequent RL cold-start training.

\begin{table}[!h]

\small
\centering
\caption{Ablation results on \bench for different learning rate and epochs in SFT.}
% \scalebox{0.8}{
\resizebox{\textwidth}{!}{
\renewcommand{\arraystretch}{1.5}
\setlength{\tabcolsep}{4pt}
\renewcommand{\arraystretch}{1.5}
\begin{tabular}{l|ccc|c|ccc|c|ccc|c}
\toprule

\multirow{2}{*}{\textbf{SFT ablations}} & \multicolumn{4}{c|}{Accuracy} & \multicolumn{4}{c|}{Relaxed Accuracy} & \multicolumn{4}{c}{PLCC}  \\  \cmidrule{2-13}
     & Visual  & Align  & Phy  & Avg  & Visual    & Align    & Phy    & Avg    & Visual & Align & Phy & Avg  \\
\midrule
Main (LR = 5e-5, 2epoch) & 43.69 & 40.88 & 34.87 & 39.81 & 90.38 & 86.97 & 83.77 & 87.04 & 56.74 & 58.24 & 44.72 & 53.23  \\
\midrule
Ablation (LR = 1e-5) & 41.60 & 38.20 & 31.20 & 37.00 & 87.80 & 81.40 & 79.40 & 82.87 & 47.37 & 45.80 & 35.40 & 42.86 \\
Ablation (LR = 2e-5) & 42.77 & 38.55 & 34.94 & 38.75 & 90.76 & 85.14 & 80.72 & 85.54 & 54.17 & 52.77 & 40.99 & 49.31 \\
Ablation (LR = 1e-4) & 41.08 & 41.48 & 37.48 & 40.01 & 88.58 & 87.38 & 81.76 & 85.91 & 53.73 & 56.94 & 42.87 & 51.18 \\
Ablation (LR = 2e-4) & 41.48 & 40.48 & 37.48 & 39.81 & 89.38 & 87.58 & 83.37 & 86.78 & 51.93 & 56.15 & 45.53 & 51.20 \\
Ablation (1epoch)  & 42.29 & 40.28 & 30.26 & 37.61 & 90.78 & 87.98 & 79.16 & 85.97 & 50.42 & 56.30 & 32.92 & 46.55 \\
Ablation (3epoch)  & 45.29 & 37.28 & 38.88 & 40.48 & 92.39 & 87.38 & 85.77 & 88.51 & 58.34 & 56.71 & 49.60 & 54.88  \\
\bottomrule
\end{tabular}
}

\label{tab:apdx_ablation_lr_epoch}
\end{table}

% =======================
\clearpage
\subsection{Ablation on inference settings}
\label{subsec:apdx_ablation_infer}
We also conduct an ablation on inference sampling rates, testing 2 fps, 4 fps, and 8 fps on \bench. Results in Table ~\ref{tab:apdx_ablation_infer} show that 2 fps achieves the best performance, which aligns with our expectation: two frames per second are sufficient to capture most quality issues for evaluation, while higher frame rates introduce redundant information and potential noise that may interfere with the model’s judgment.

\begin{table}[h]

\small
\centering
\caption{Ablation results on \bench for different inference configurations.}
\label{tab:apdx_ablation_infer}
% \scalebox{0.8}{
\resizebox{\textwidth}{!}{
\setlength{\tabcolsep}{6pt}
\renewcommand{\arraystretch}{1.5}
\begin{tabular}{c|ccc|c|ccc|c|ccc|c}
\toprule

Inference & \multicolumn{4}{c|}{Accuracy} & \multicolumn{4}{c|}{Relaxed Accuracy} & \multicolumn{4}{c}{PLCC}  \\  \cmidrule{2-13}
  Sampling fps      & Visual  & Align  & Phy  & Avg  & Visual    & Align    & Phy    & Avg    & Visual & Align & Phy & Avg  \\
\midrule
 2fps & \textbf{50.10} & 43.88 & \textbf{39.08} & \textbf{44.35} & 92.99 & \textbf{91.38} & \textbf{87.98} & \textbf{90.78} & \textbf{60.13} & \textbf{62.60} & \textbf{52.73} & \textbf{60.37} \\
 
 4fps & 46.80 & \textbf{44.20} & 38.28 & 43.09 & 90.00 & 87.60 & 84.20 & 87.27 & 60.13 & 57.53 & 43.55 & 53.74 \\
 
 8fps & 41.67 & 40.77 & 37.61 & 40.02 & 85.81 & 88.96 & 83.78 & 86.18 & 56.28 & 58.27 & 41.86 & 52.14 \\
\bottomrule 
\end{tabular}
}

\end{table}

% =======================
\clearpage
\subsection{Ablation on RL training steps}
\label{subsec:apdx_rl_steps}

We evaluated multiple intermediate checkpoints during RL training. Considering three evaluation metrics jointly, performance peaked around 300 steps. Beyond this point, scores on \bench showed a clear decline., as shown in Table~\ref{tab:apdx_ablation_rl_steps}. Therefore, for all main experiments, we report results based on the 300-step checkpoint.

\begin{table}[ht]

\small
\centering
\caption{Ablation on RL training steps. Accuracy and correlation between model answer and human score on \bench. \textit{Relaxed Accuracy} counts cases where the prediction differs from the ground truth by at most one point.}
\label{tab:apdx_ablation_rl_steps}
% \scalebox{0.84}{
\resizebox{\textwidth}{!}{
\renewcommand{\arraystretch}{1.35}
\setlength{\tabcolsep}{5pt}
\begin{tabular}{l|ccc|c|ccc|c|ccc|c}
\toprule

\multirow{2}{*}{RL steps} & \multicolumn{4}{c|}{Accuracy} & \multicolumn{4}{c|}{Relaxed Accuracy} & \multicolumn{4}{c}{PLCC}  \\  \cmidrule{2-13}
 & Visual  & Align  & Phy  & Avg  
 & Visual  & Align  & Phy  & Avg    
 & Visual  & Align  & Phy  & Avg  \\ 
 \midrule
200 & 50.50 & 42.89 & 39.28 & 44.22 & 92.79 & 91.59 & 87.80 & 90.73 & 65.14 & 62.95 & 57.60 & 61.90 \\
300 & 50.10 & 43.88 & 39.08 & 44.35 & 92.99 & 91.38 & 87.98 & 90.78 & 65.78 & 62.60 & 52.73 & 60.37 \\
400 & 46.20 & 43.80 & 36.00 & 42.00 & 92.80 & 90.20 & 85.40 & 89.47 & 64.57 & 58.87 & 44.61 & 61.72 \\
500 & 47.60 & 45.80 & 40.00 & 44.47 & 90.20 & 91.40 & 87.80 & 89.80 & 61.57 & 60.59 & 52.10 & 58.09 \\
600 & 50.60 & 43.40 & 41.40 & 45.13 & 91.40 & 89.90 & 87.20 & 89.50 & 62.89 & 56.49 & 51.62 & 57.00 \\
700 & 48.00 & 45.20 & 38.00 & 43.73 & 90.80 & 88.20 & 87.60 & 88.87 & 64.28 & 57.07 & 49.43 & 56.93 \\
833 & 45.00 & 45.60 & 37.80 & 42.80 & 91.40 & 89.60 & 85.60 & 88.87 & 64.68 & 59.19 & 46.04 & 56.64 \\

% 300  & \textbf{50.10} & \textbf{43.88} & \textbf{39.08} & \textbf{44.35} & \textbf{92.99} & \textbf{91.38} & \textbf{87.98} & \textbf{90.78} & \textbf{60.13} & \textbf{62.60} & \underline{52.73} & \textbf{60.37}   \\

\bottomrule
\end{tabular}
}
\vspace{-4pt}

\end{table}

% ======================================
\clearpage
\section{Case Studies}
\label{sec:apdx_case_study}

\begin{figure}[ht]
    \centering
    \includegraphics[width=0.75\linewidth]{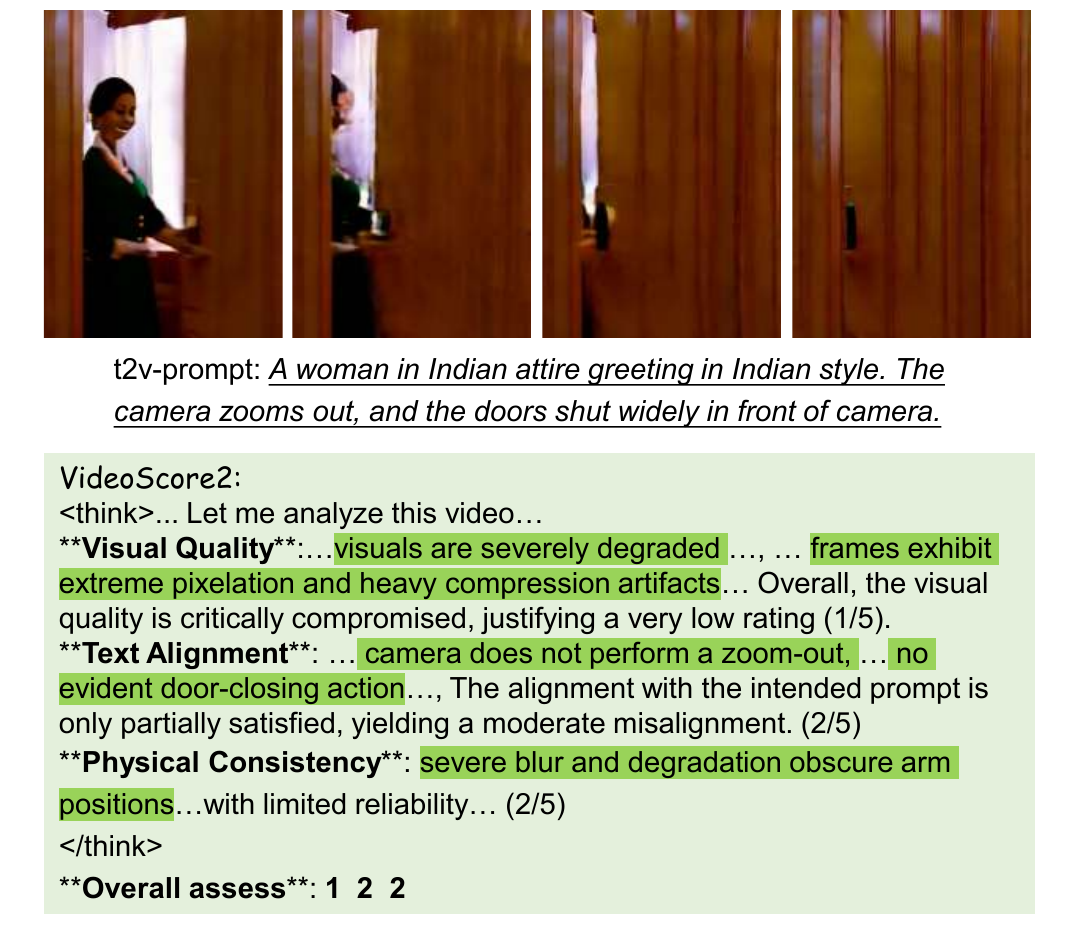}
    \label{fig:apdx_case1}
    \caption{Case study of \score evaluation (1)}
\end{figure}

\begin{figure}[ht]
    \centering
    \includegraphics[width=0.75\linewidth]{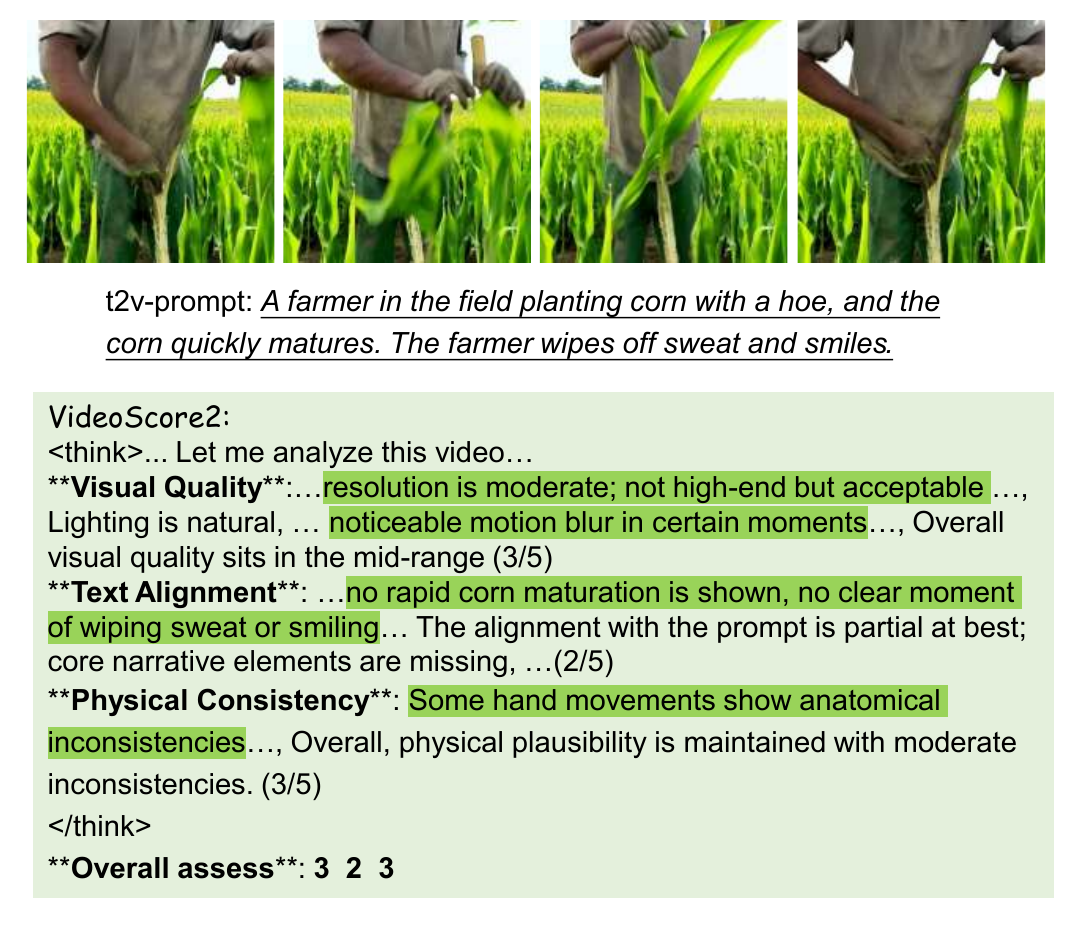}
    \label{fig:apdx_case2}
    \caption{Case study of \score evaluation (2)}
\end{figure}

\begin{figure}[ht]
    \centering
    \includegraphics[width=0.75\linewidth]{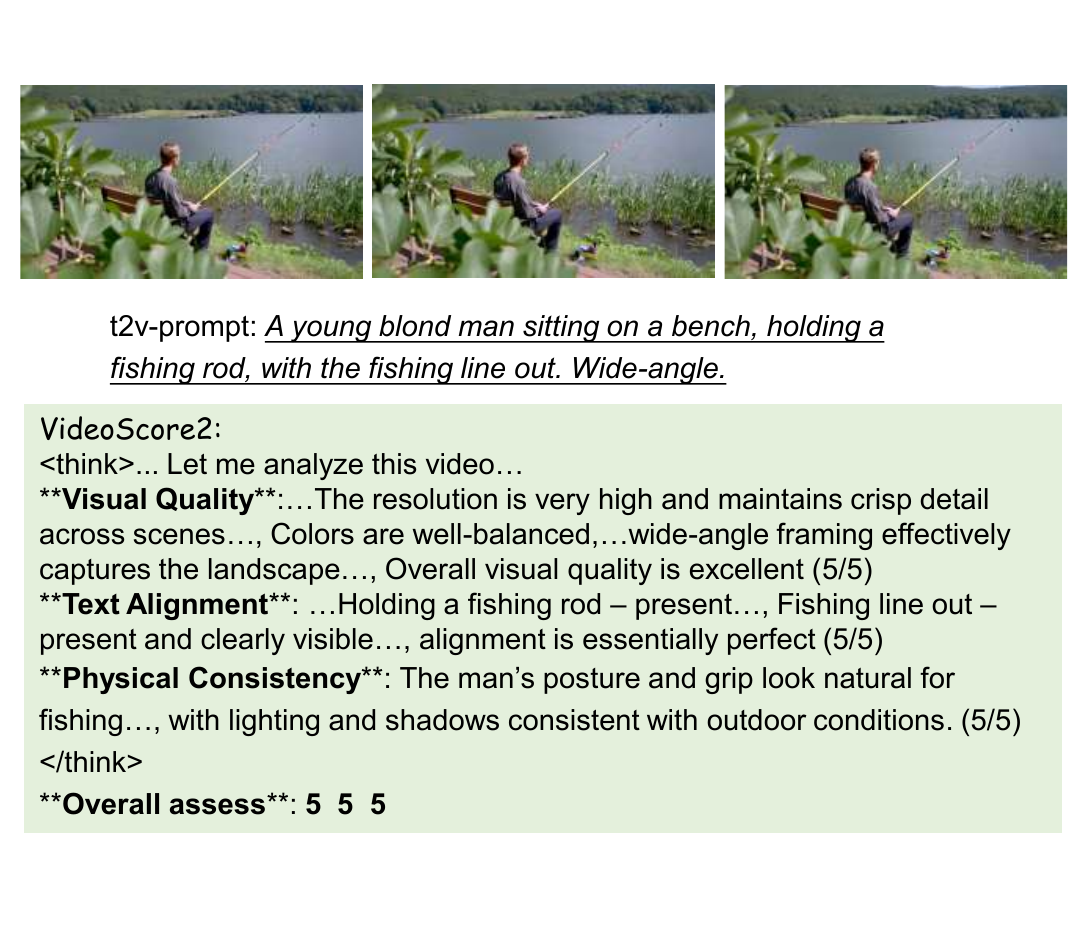}
    \label{fig:apdx_case3}
    \caption{Case study of \score evaluation (3)}
\end{figure}

% ======================================
\clearpage
\section{The Use of Large Language Models}
\label{sec:apdx_use_llms}

Large language models (LLMs), including GPT-5, Gemini-2.5-Pro were used in the preparation of this paper. Their role was limited to supporting writing by suggesting phrasing alternatives, correcting grammar, and improving readability. All technical content, experimental design, analysis, and conclusions were created and verified by the authors.

\clearpage

\end{document}